\documentclass[10pt,twocolumn,letterpaper]{article}

\usepackage{cvpr}              

\usepackage{graphicx}
\usepackage{amsmath}
\usepackage{amssymb}
\usepackage{booktabs}

\usepackage{algorithm}
\usepackage{algorithmic}
\usepackage{multirow}

\usepackage[pagebackref,breaklinks,colorlinks]{hyperref}

\usepackage[capitalize]{cleveref}
\crefname{section}{Sec.}{Secs.}
\Crefname{section}{Section}{Sections}
\Crefname{table}{Table}{Tables}
\crefname{table}{Tab.}{Tabs.}

\begin{document}
	
	\title{Federated Learning with Position-Aware Neurons}
	
	\author{
		Xin-Chun Li$^{1}$, Yi-Chu Xu$^{1}$, Shaoming Song$^{2}$, Bingshuai Li$^{2}$, Yinchuan Li$^{2}$, \\
		Yunfeng Shao$^{2}$, De-Chuan Zhan$^{1}$ \\  
		$^{1}$State Key Laboratory for Novel Software Technology, Nanjing University \\ $^{2}$Huawei Noah’s Ark Lab\\
		{\tt\small 
			\{lixc, xuyc\}@lamda.nju.edu.cn, zhandc@nju.edu.cn
		} \\
		{\tt\small 
			\{shaoming.song, libingshuai, liyinchuan, shaoyunfeng\}@huawei.com
		}
	}
	\maketitle
	
	\begin{abstract}
		Federated Learning (FL) fuses collaborative models from local nodes without centralizing users' data. The permutation invariance property of neural networks and the non-i.i.d. data across clients make the locally updated parameters imprecisely aligned, disabling the coordinate-based parameter averaging. Traditional neurons do not explicitly consider position information. Hence, we propose Position-Aware Neurons (PANs) as an alternative, fusing position-related values (i.e., position encodings) into neuron outputs. PANs couple themselves to their positions and minimize the possibility of dislocation, even updating on heterogeneous data. We turn on/off PANs to disable/enable the permutation invariance property of neural networks. PANs are tightly coupled with positions when applied to FL, making parameters across clients pre-aligned and facilitating coordinate-based parameter averaging. PANs are algorithm-agnostic and could universally improve existing FL algorithms. Furthermore, ``FL with PANs" is simple to implement and computationally friendly.
	\end{abstract}
	
	
	\section{Introduction}
	\label{sec:intro}
	
	
	Federated Learning (FL)~\cite{Fed-Advances,Fed-Concept} generates a global model via collaborating with isolated clients for privacy protection and efficient distributed training, generally following the parameter server architecture~\cite{ParameterServer,LargeScaleNet}. Clients update models on their devices using private data, and the server periodically averages these models for multiple communication rounds~\cite{FedAvg}. The whole process does not transmit users' data and meets the basic privacy requirements.
	
	
	\begin{figure}[t]
		\centering
		\includegraphics[width=\linewidth]{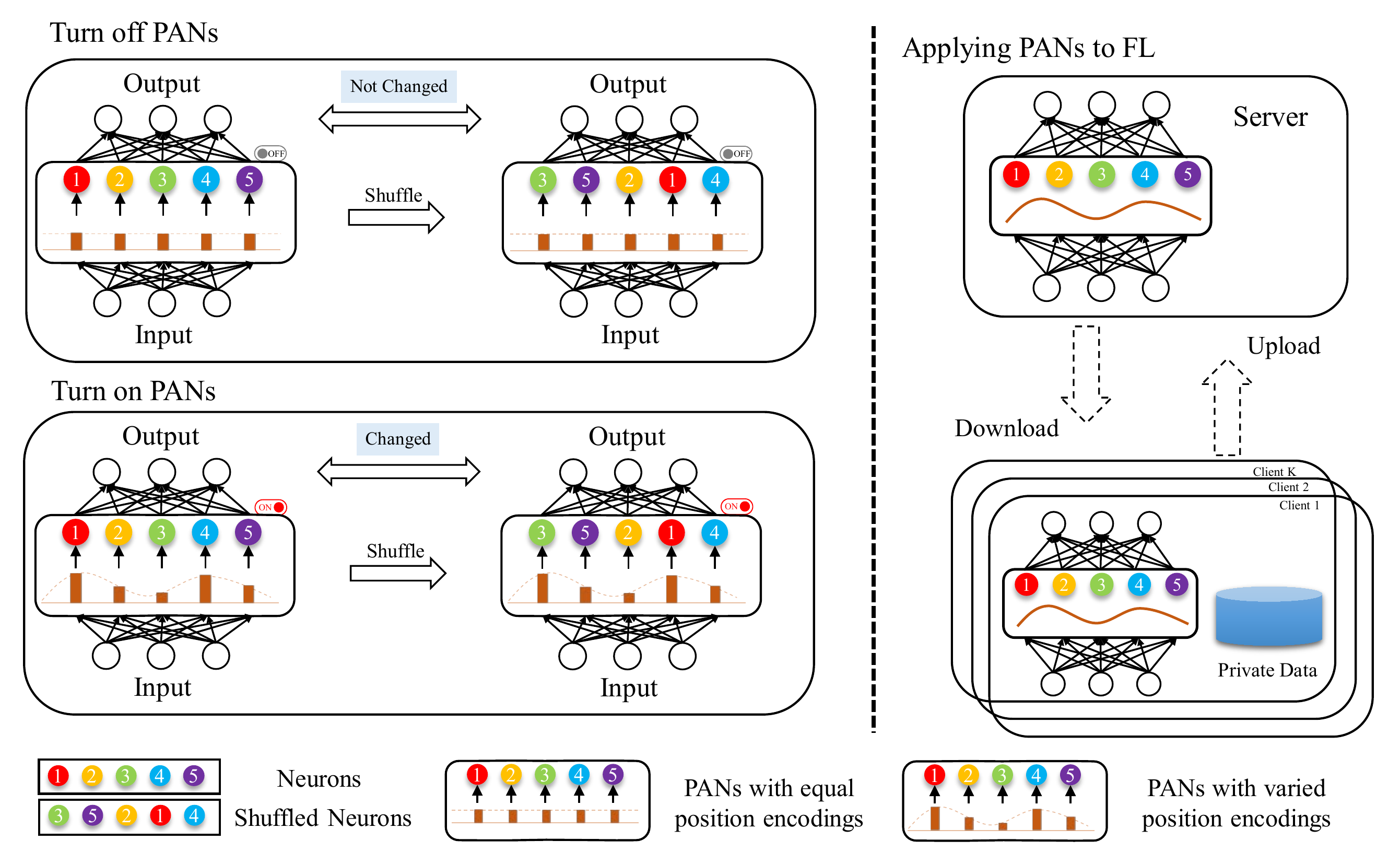}
		\caption{{\bf Left}: Position-Aware Neurons (PANs). We fuse equal/varied position encodings to neurons' outputs, PANs are turned off/on, and the shuffled networks make the same/different predictions, i.e., the permutation invariance property is enabled/disabled. {\bf Right}: applying PANs to FL. Neurons are coupled with their positions for pre-alignment.}
		\label{fig:teaser}
	\end{figure}
	
	Represented by FedAvg~\cite{FedAvg}, many FL algorithms aggregate local parameters via a simple coordinate-based averaging~\cite{FedProx,FedRS,MOON,FedPHP} These algorithms have two kinds of drawbacks. First, as traditional neurons are unaware of their positions, neural networks have the permutation invariance property, implying that hidden neurons could be dislocated during training without affecting the local performances. Second, the samples across clients are non-independent and identically distributed (non-i.i.d.)~\cite{NonIID-Quag}, which could exacerbate the permutation of neural networks during local training, making local models misaligned and leading to weight divergence~\cite{Fed-NonIID-Data}. These reasons degrade the performance of coordinate-based parameter averaging.
	
	
	Recently, a series of works utilize various matching techniques to align neurons, such as Bayesian nonparametric learning~\cite{BFNM,SMAPM,FedMA} and optimal transport~\cite{Barycenter,OTFusion}. First, these methods are too complex to implement. Second, they solve the misalignment problem after finishing local updates and hence belong to post-processing strategies that need additional computation budgets. Fed$^2$~\cite{Fed2} pioneers a novel aspect via designing feature-oriented model structures following a pre-aligned manner. However, it has to carefully customize the network architecture and only stays at the group level of pre-alignment. By contrast, we explore a more straightforward and general technique to pre-align neurons during local training procedures.
	
	Our work mainly focuses on solving the non-i.i.d. challenge in FL, more specifically, seeking solutions via limiting the permutation invariance property of neural networks. We first summarize the above analysis: {\it the permutation invariance property of neural networks leads to neuron misalignment across local models. The more heterogeneous the data, the more serious the misalignment is}. Hence, our motivation is intuitive: {\it could we design a switch to control the permutation invariance property of neuron networks?} We propose {\bf Position-Aware Neurons (PANs)} as the solution, which couple neurons with their positions. Specifically, for each neuron (channel for ConvNet~\cite{AlexNet,VGG,ResNet}), we add or multiply a position-related value (i.e., position encoding) to its output. We introduce a hyper-parameter to turn on/off the PANs, and correspondingly, to disable/enable the permutation invariance property of neural networks. PANs bind neurons in their positions, implicitly pre-aligning neurons across clients even faced with non-i.i.d. data. From another aspect, PANs could keep some consistent ingredients in the forward and backward pass across local models, which could reduce the weight divergence. Overall, appropriate PANs facilitate the coordinate-based parameter averaging in FL. Replacing traditional neurons with PANs is simple to implement and computationally friendly, which is universal to various FL algorithms. Contributions can be briefed as: {\it(1) proposing PANs to disable/enable the permutation invariance property of deep networks; (2) applying PANs to FL, which binds neurons in positions and pre-aligns parameters for better coordinate-wise parameter averaging.} 
	
	\section{Related Works}
	\label{sec:relate}
	\noindent \textbf{FL with Non-I.I.D. Data:} Existing works solve the non-i.i.d. data problem in FL from various aspects. \cite{Fed-NonIID-Data} points out the weight divergence phenomenon in FL and use shared data to decrease the divergence. FedProx~\cite{FedProx} takes a proximal term during local training as regularization. FedOpt~\cite{FedOpt} considers updating the global model via momentum or adaptive optimizers (e.g., Adam~\cite{Adam}, Yogi~\cite{Yogi}) instead of simple parameter averaging. Scaffold~\cite{Scaffold} introduces control variates to rectify the local update directions and mitigates the influences of client drift. MOON~\cite{MOON} utilizes model contrastive learning to reduce the distance between local and global models. Some other works utilize similar techniques including dynamic regularization~\cite{FedDyn}, ensemble distillation~\cite{FedDF,OnlineDistill}, etc. We take several representative FL algorithms and use PANs to improve them.
	
	\noindent \textbf{FL with Permutation Invariance Property:} The permutation invariance of neuron networks could lead to neuron misalignment. PFNM~\cite{BFNM} matches local nodes' parameters via Beta-Bernoulli process~\cite{BBP} and Indian Buffet Process~\cite{IBP}, formulating an optimal assignment problem and solving it via Hungarian algorithm~\cite{Hungarian}. SPAHM~\cite{SMAPM} applies the same procedure to aggregate Gaussian topic models, hidden Markov models, and so on. FedMA~\cite{FedMA} points out PFNM does not apply to large-scale networks and proposes a layer-wise matching method. \cite{OTFusion} utilizes optimal transport~\cite{Barycenter} to fuse models with different initializations. These methods are all post-processing ones that need additional computation costs. Fed$^2$ is recently proposed to align features during local training via separating features into different groups. However, it needs to carefully design the architectures. Differently, we take a more fine-grained alignment of neurons rather than network groups, and we will show our method is more general.
	
	\noindent \textbf{Position Encoding:} Position encoding is popular in sequence learning architectures, e.g., ConvS2S~\cite{ConvS2S} and transformer~\cite{Transformer}, etc. These architectures take position encodings to consider the order information. Relative position encoding~\cite{RPE} is more applicable to sequences with various lengths. Some other studies are devoted to interpreting what position encodings learn~\cite{PEInBERT,WhatPELearn}. Another interesting work is applying position encodings instead of zero-padding to GAN~\cite{PEGAN} as spatial inductive bias. Differently, we resort to position encodings to bind neurons in their positions in FL. Furthermore, these works only consider position encodings at the input layer, while we couple them with neurons.
	
	\section{Position-Aware Neurons}
	In this section, we investigate the permutation invariance of neural networks and introduce PANs to control it.
	
	
	\subsection{Permutation Invariance Property}
	Assume an MLP network has $L+1$ layers (containing input and output layer), and each layer contains $J_{l}$ neurons, where $l \in \{0,1,\cdots,L\}$ is the layer index. $J_0$ and $J_L$ are input and output dimensions. We denote the parameters of each layer as the weight matrix $W_{l} \in \mathcal{R}^{J_{l} \times J_{l-1}}$ and the bias vector $b_{l} \in \mathcal{R}^{J_{l}}$, $l\in \{1,2,\cdots,L\}$. The input layer does not have parameters. We use $h_l \in \mathcal{R}^{J_l}$ as the activations of the $l$th layer. We have $h_{l}=f_{l}(W_lh_{l-1} + b_l)$, where $f_l(\cdot)$ is the element-wise activation function, e.g., ReLU~\cite{relu}. $f_L(x)=x$ denotes no activation function in the output layer. Sometimes, we use $y=v^Tf(Wx+b)$ to represent a network with only one hidden layer and the output dimension is one (called as MLP0), where $x\in \mathcal{R}^{J_0}$, $W\in \mathcal{R}^{J \times J_0}$, $b\in \mathcal{R}^J$, $v\in \mathcal{R}^J$. We use $\Pi \in \{0,1\}^{J\times J}$ as a permutation matrix that satisfies $\sum_{j}\Pi_{\cdot,j}=1$ and $\sum_{j}\Pi_{j,\cdot}=1$. Easily, we have some properties: $\Pi^T\Pi=I$, $\Pi a + \Pi b = \Pi(a + b)$, $\Pi a \odot \Pi b = \Pi (a \odot b)$, where $I$ is the identity matrix and $\odot$ denotes Hadamard product. If $f(\cdot)$ is an element-wise function, $f(\Pi x) = \Pi f(x)$.
	
	For MLP0, we have $y=(\Pi v)^Tf(\Pi Wx+\Pi b)=v^Tf(Wx+b)$, implying that if we permute the parameters properly, the output of a certain neural network does not change, i.e., the {\bf permutation invariance property}. Extending it to MLP, the layer-wise permutation process is
	\begin{equation}
		h_l = f_l(\Pi_l W_l \Pi_{l-1}^T h_{l-1} + \Pi_l b_l), \label{eq:permute}
	\end{equation}
	where $\Pi_0=I$ and $\Pi_L=I$, meaning that the input and output layers are not shuffled. 
	For ConvNet~\cite{AlexNet,VGG}, we take convolution kernels as basic units. The convolution parameters could be denoted as $W_{l}\in \mathcal{R}^{C_{l}\times w_{l} \times h_{l} \times C_{l-1}}$, where the four dimensions denote the number of output/input channels ($C_l$, $C_{l-1}$) and the kernel size ($w_l$, $h_l$). The permutation could be similarly applied as $\Pi_{l}W_l\Pi_{l-1}^T$. For ResNet~\cite{ResNet}, we use $h_{l}=f_l(\Pi_l W_l \Pi_{l-1}^T h_l) + \Pi_l M_l \Pi_{l-1}^T h_l$ to permute all parameters in a basic block including the shortcut (if shortcut is not used, $M_l=I$).
	
	
	\subsection{Position-Aware Neurons} \label{sec:pan}
	The essential reason for the permutation invariance of neural networks is that neurons have nothing to do with their positions. Hence, an intuitive improvement is fusing position-related values (position encodings) to neurons. We propose {\bf Position-Aware Neurons (PANs)}, adding or multiplying position encodings to neurons' outputs, i.e.,
	\begin{align}
		\text{PAN}_{+}&:~h_l = f_l(W_l h_{l-1} + b_l + \underline{e_l}), \label{eq:add-pan} \\
		\text{PAN}_{\circ}&:~h_l = f_l((W_l h_{l-1} + b_l) \odot \underline{e_l}), \label{eq:mul-pan}
	\end{align}
	where $e_l$ denotes position encodings that are only related to positions and not learnable. We use ``$\text{PAN}_{+}$" and ``$\text{PAN}_{\circ}$" to represent additive and multiplicative PANs, respectively. We use sinusoidal functions to generate $e_l$ as commonly used in previous position encoding works~\cite{Transformer}, i.e., 
	\begin{align}
		\text{PAN}_{+}&:~e_{l,j}= A \sin \left( 2\pi T j/J \right) \in [-A, A], \label{eq:add-pe} \\ 
		\text{PAN}_{\circ}&:~e_{l,j}= 1 + A \sin \left( 2\pi T j/J \right) \in [1-A, 1+A], \label{eq:mul-pe}
	\end{align}
	where $T$ and $A$ respectively denotes the period and amplitude of position encodings, and $j\in \{0, 1,\cdots,J-1\}$ is the position index of a neuron. {\it For ConvNet, we assign position encodings for each channel, and $j$ is the channel index. Notably, if we take $T\rightarrow 0$ or $A=0$, PANs degenerate into normal neurons. In practice, we only apply PANs to the hidden layers, while the input and output layers remain unchanged, i.e., $l \in \{1,2,\cdots,L-1\}$ for $e_l$.} With PANs, the permutation process in Eq.~\ref{eq:permute} could be reformulated as
	\begin{align}
		\text{PAN}_{+}&:~h_{l,\text{sf}}= f_l(\Pi_l W_l \Pi_{l-1}^T h_{l-1,\text{sf}} + \Pi_l b_l + \underline{e_l}), \label{eq:permute-add-pe} \\
		\text{PAN}_{\circ}&:~h_{l,\text{sf}} = f_l((\Pi_l W_l \Pi_{l-1}^T h_{l-1,\text{sf}} + \Pi_l b_l) \odot \underline{e_l}), \label{eq:permute-mul-pe}
	\end{align}
	where the subscript ``$\text{sf}$" denotes ``shuffled" (or permuted). To measure the output change after shuffling, we define the {\it shuffle error} as:
	\begin{equation}
		\text{Err}(A, T, \{\Pi_{l}\}_{l=0}^L)=\lVert h_{L,\text{sf}} - h_{L} \rVert / J_L, \label{eq:sf-error}
	\end{equation}
	and this error on MLP0 without considering bias (i.e., $y=v^T f(Wx + e)$) is
	\begin{align}
		\nonumber
		\text{PAN}_{+}&:~\text{Err}(A, T, \Pi) \\
		\nonumber
		&= \lvert y_{\text{sf}} - y \rvert \\ \nonumber
		&= \lvert (\Pi v)^T f(\Pi Wx + e) - v^T f(Wx + e) \rvert \\ \nonumber
		&= \lvert (\Pi v)^T f(\Pi Wx + e) - (\Pi v)^T f(\Pi Wx + \Pi e) \rvert \\
		&\approx \lvert (\Pi e - e)^T \frac{\partial{y_{\text{sf}}}}{\partial e}\rvert,
		\label{eq:sf-error-mlp0}
	\end{align}
	where we take $y_{\text{sf}}=(\Pi v)^T f(\Pi Wx + e)$ as the function of $e$ and take Taylor expansion as an approximation. Obviously, shuffle error is closely related to the strength of permutation, i.e., $\Pi - I$. For example, if $\Pi=I$, the network is not shuffled and the outputs are kept unchanged. Then, if we take equal values as position encodings, i.e., $e_{j}=e_{i}, \forall i, j$, the output also does not change because $\Pi e = e$. This can be obtained via taking $\alpha=0$ or $T \rightarrow 0$. If we take a larger $T$ (e.g., 1) and larger $\alpha$ (e.g., 0.05), $\text{Err}$ is generally non-zero because $\Pi e \neq e$. The error of multiplicative PANs is similar. {\it We abstract PANs as a switch: if we take equal/varied position encodings, PANs are turned off/on, and hence the network keeps/loses the permutation invariance property (i.e., the same/different outputs after permutation)}. As illustrated at the left of Fig.~\ref{fig:teaser}, the five neurons of a certain hidden layer are shuffled while the position encodings they are going to add/multiply are not shuffled, and the outputs will change with PANs turned on.
	
	Furthermore, are there any essential differences between additive and multiplicative PANs, and how much influence do they have on the shuffle error? In Eq.~\ref{eq:sf-error-mlp0}, the shuffle error is partially determined by $\partial{y_{\text{sf}}}/\partial{e}$, and we extent this gradient to MLP with multiple layers. We assume all layers have the same number of neurons (i.e., $J_l=J, \forall l$) and take the same position encodings (i.e., $e_l=e \in \mathcal{R}^J, \forall l$). We denote $s_{l,\text{sf}}=\Pi_l W_l \Pi_{l-1}^T h_{l-1,\text{sf}} + \Pi_l b_l$ and obtain the recursive gradient expressions:
	\begin{align}
		\text{PAN}_{+}&:\frac{\partial{h_{l,\text{sf}}}}{{\partial e}} = \text{D}(f_{l}^{\prime})\left(\frac{\partial{s_{l,\text{sf}}}}{\partial{h_{l-1,\text{sf}}}}\frac{\partial{h_{l-1,\text{sf}}}}{\partial{e}} + I \right), \label{eq:grad-add-pe} \\
		\text{PAN}_{\circ}&:\frac{\partial{h_{l,\text{sf}}}}{{\partial e}} = \text{D}(f_{l}^{\prime}) \bigg(\frac{\partial{s_{l,\text{sf}}}}{\partial{h_{l-1,\text{sf}}}}\frac{\partial{h_{l-1,\text{sf}}}}{\partial{e}} \odot [e]^{J} \nonumber \\
		&~~~~~~~~~~~~~~~~~~~~~~~~~~~~~~~~~~~~~~~~~~~~~~~~~~~~+ \text{D}(s_{l,\text{sf}}) \bigg), \label{eq:grad-mul-pe}
	\end{align}
	where $\text{D}(\cdot)$ transforms a vector to a diagonal matrix and $[\cdot]^J$ repeats a vector $J$ times to obtain a matrix. $f_{l}^{\prime}$ denotes the gradient of activation functions, whose element is 0 or 1 in ReLU. If we expand Eq.~\ref{eq:grad-add-pe} and Eq.~\ref{eq:grad-mul-pe} correspondingly, we will find that the gradient $\frac{\partial{h_{L, \text{sf}}}}{\partial e}$ of additive PANs does not explicitly rely on $e$. However, for the multiplicative one, $\frac{\partial{h_{l,\text{sf}}}}{{\partial e}}$ is relevant to $\frac{\partial{h_{l-1,\text{sf}}}}{\partial{e}}$ and $[e]^{J}$, which could lead to a polynomial term $A^{L-1}$ (resulted from $[e]^J \odot \cdots \odot [e]^J$, informally). Hence, we conclude: {\it taking PANs as a switch could control the permutation invariance property of neural networks. The designed multiplicative PANs will make this switch more sensitive.}
	
	

	\section{FL with PANs}
	In this section, we briefly introduce FedAvg~\cite{FedAvg} and analyze the effects of PANs when applied to FL.
	
	\subsection{FedAvg}
	Suppose we have a server and $K$ clients with various data distributions. FedAvg first initializes a global model $\theta_{0}$ on the server. Then, a small fraction (i.e. $R \in [0, 1]$) of clients $S_t$ download the global model and update it on their local data for $E$ epochs, and then upload the updated model $\theta_{0}^{(k)}$ to the server. Then, the server takes a coordinate-based parameter averaging, i.e., $\theta_{1} \leftarrow \frac{1}{|S_t|} \sum_{k\in S_t} \theta_{0}^{(k)}$. Next, $\theta_{1}$ will be sent down for a new communication round. This will be repeated for $H$ communication rounds. Because the parameters could be misaligned during local training, some works~\cite{BFNM,FedMA,SMAPM} are devoted to finding the correspondences between clients' uploaded neurons for better aggregation. For example, the parameters $W_{l}^{(1)}$ and $W_{l}^{(2)}$ may be misaligned, and we should search for proper matrices to match them, i.e., $\frac{1}{2}(W_{l}^{(1)} + M_{l}W_{l}^{(2)}M_{l-1}^T)$, rather than $\frac{1}{2}(W_{l}^{(1)} + W_{l}^{(2)})$~\cite{OTFusion}. However, searching for appropriate $M_{\{l,l-1\}}$ is challenging. Generally, these works require additional data to search for proper alignment. In addition, the matching process has typically to solve complex optimization problems, such as optimal transport or optimal assignment, leading to additional computational overhead. An intuitive question is: {\it could we pre-align the neurons during local training instead of post-matching?}

	
	\subsection{Applying PANs to FL} \label{sec:pan-in-fl}
	Replacing traditional neurons with PANs in FL is straightforward to implement. Why does such a subtle improvement help? We heuristically expect PANs in FL could bring such effects: {\it PANs could limit the dislocation of neurons since the disturbance of them will bring significant changes to the outputs of the neural network and lead to higher training errors and fluctuations}. Theoretically, the forward pass on the $k$th client with PANs is as follows:
	\begin{align}
		\text{PAN}_{+}:~h_l^{(k)} &= f_l^{(k)}(W_l^{(k)} h_{l-1}^{(k)} + b_l^{(k)} + \underline{e_l}), \label{eq:fl-add-pan} \\
		\text{PAN}_{\circ}:~h_l^{(k)} &= f_l^{(k)}((W_l^{(k)} h_{l-1}^{(k)} + b_l^{(k)}) \odot \underline{e_l}). \label{eq:fl-mul-pan}
	\end{align}
	Notably, the position encodings are commonly utilized across clients, i.e., the forward pass across local clients share some consistent information. Then, the parameters' gradient of Eq.~\ref{eq:fl-add-pan} and Eq.~\ref{eq:fl-mul-pan} can be calculated by:
	\begin{align}
		\text{PAN}_{+}&:~\partial{h_l^{(k)}}/\partial{b_l^{(k)}} = \text{D}({f_l^{(k)}}^{\prime}),
		\label{eq:grad-fl-add-pan} \\
		\text{PAN}_{\circ}&:~\partial{h_l^{(k)}}/\partial{b_l^{(k)}} = \text{D}({f_l^{(k)}}^{\prime})\text{D}(e_l), \label{eq:grad-fl-mul-pan}
	\end{align}
	where we only give the gradient of bias for simplification. The gradients of multiplicative PANs directly contain the same position information across clients (e.g., $e_l$) in spite of various data distributions (e.g., $h_{l-1}^{(k)}$). For the additive ones, the impact of $e_l$ is implicit because ${f_{l}^{(k)}}^{\prime}$ is related to $e_l$, but nevertheless, the effect is not significant as multiplicative ones. {\it Overall, $e_l$ could regularize and rectify local gradient directions, keeping some ingredients consistent during backward propagation.} As an extreme case, if $A$ in $e_l$ is very large, the gradients in Eq.~\ref{eq:grad-fl-add-pan} and Eq.~\ref{eq:grad-fl-mul-pan} will tend to be the same, mitigating the weight divergence completely. However, setting $e_l$ too large will make the neural network difficult to train and the data information is completely covered, so the strength of $e_l$ (i.e., $A$) is a tradeoff.
	
	\begin{figure}[t]
		\centering
		\includegraphics[width=\linewidth]{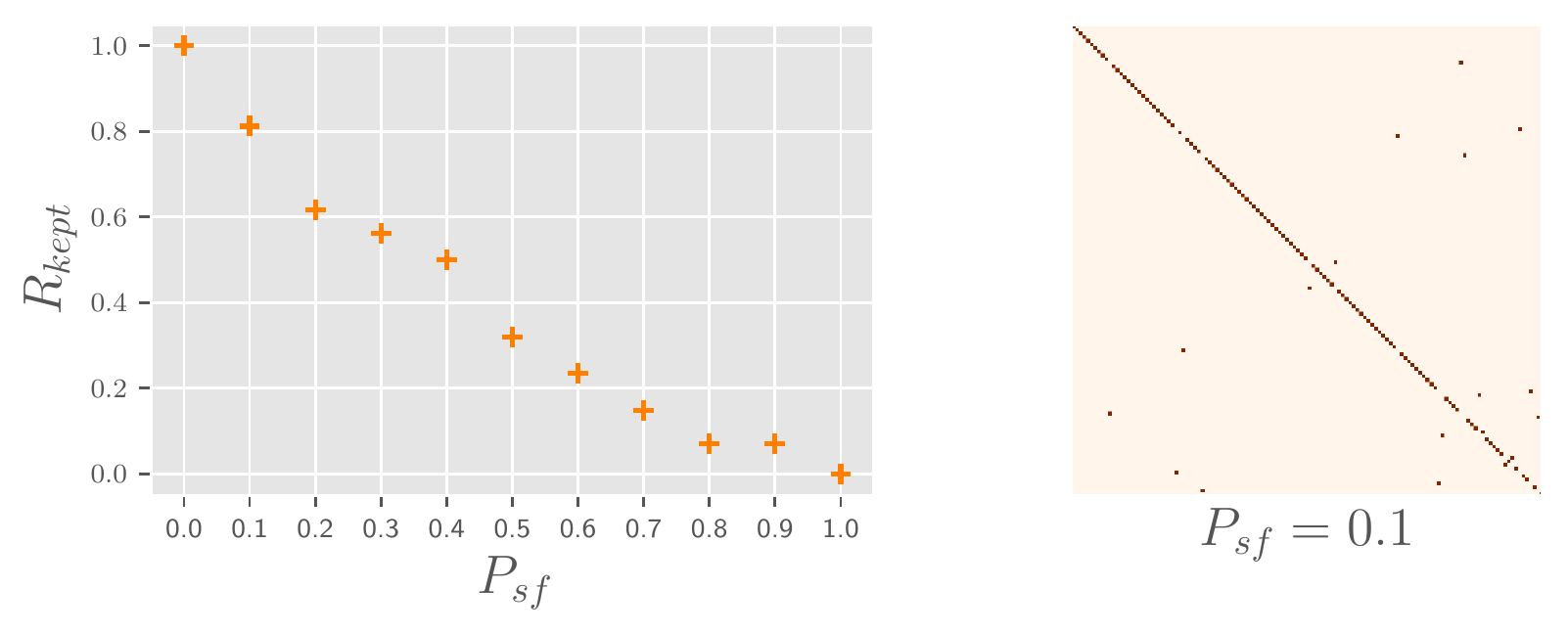}
		\caption{{\bf Left}: how much neurons are not shuffled with various $P_{\text{sf}}$. {\bf Right}: a permutation matrix demo with $P_{\text{sf}}=0.1$.}
		\label{fig:sf-prob}
	\end{figure}
	
	\section{Experiments}
	We study how much influence the proposed PANs have on both centralized training and decentralized training (i.e., FL). The datasets used are Mnist~\cite{mnist}, FeMnist~\cite{LEAF}, SVHN~\cite{Svhn}, GTSRB~\cite{GTSRB}, Cifar10/100~\cite{cifar}, and Cinic10~\cite{Cinic10}. FeMnist is recommended by LEAF~\cite{LEAF} and FedScale~\cite{FedScale}. We use MLP for Mnist/FeMnist, VGG~\cite{VGG} for SVHN/GTSRB/Cifar10, ResNet20~\cite{ResNet} for Cifar100/Cinic10 by default if without more declarations. We sometimes take VGG9 used in previous FL works~\cite{FedMA,FedDF,Fed2}. For centralized training, we use the provided training and test set correspondingly. For FL, we split the training set according to Dirichlet distributions, where $\text{Dir}(\alpha)$ controls the non-i.i.d. level. Smaller $\alpha$ leads to more non-i.i.d. cases. For each FL scene, we report several key hyper-parameters: number of clients $K$, client participation ratio $R$, number of local training epochs $E$, Dirichlet alpha $\alpha$, number of communication rounds $H$. For PANs, we report $T$ and $A$. {\it With $A=0.0$, we turn off PANs, i.e., using traditional neurons or the baselines; with $A>0.0$, we turn on PANs. We {\bf leave PANs turned on by default} if with no mention of the state on/off or the value of $A$}. Details of datasets, networks and training are presented in Supp.
	
	\subsection{Centralized Training}
	\paragraph{Shuffle Test:} We first propose a procedure to measure the degree of permutation invariance of a certain neural network, that is, how large the shuffle error in Eq.~\ref{eq:sf-error} is after shuffling the neurons. We name this procedure {\it shuffle test}. Given a neural network and a batch of data, we first obtain the outputs. Then, we shuffle the neurons of hidden layers. The shuffle process is shown in Supp, where $P_{\text{sf}}$ controls the disorder level of the constructed permutation matrices. Then we could get the outputs after shuffling and then calculate the shuffle error. We vary $P_{\text{sf}}$ in $[0,1]$ and plot the ratio of permutation matrices' diagonal ones (i.e., how much neurons are not shuffled). We denote this ratio as $R_{\text{kept}}$ and plot them in Fig.~\ref{fig:sf-prob} (average of 10 experiments), where we also show a generated permutation matrix with $P_{\text{sf}}=0.1$.
	
	\begin{figure}[t]
		\centering
		\includegraphics[width=\linewidth]{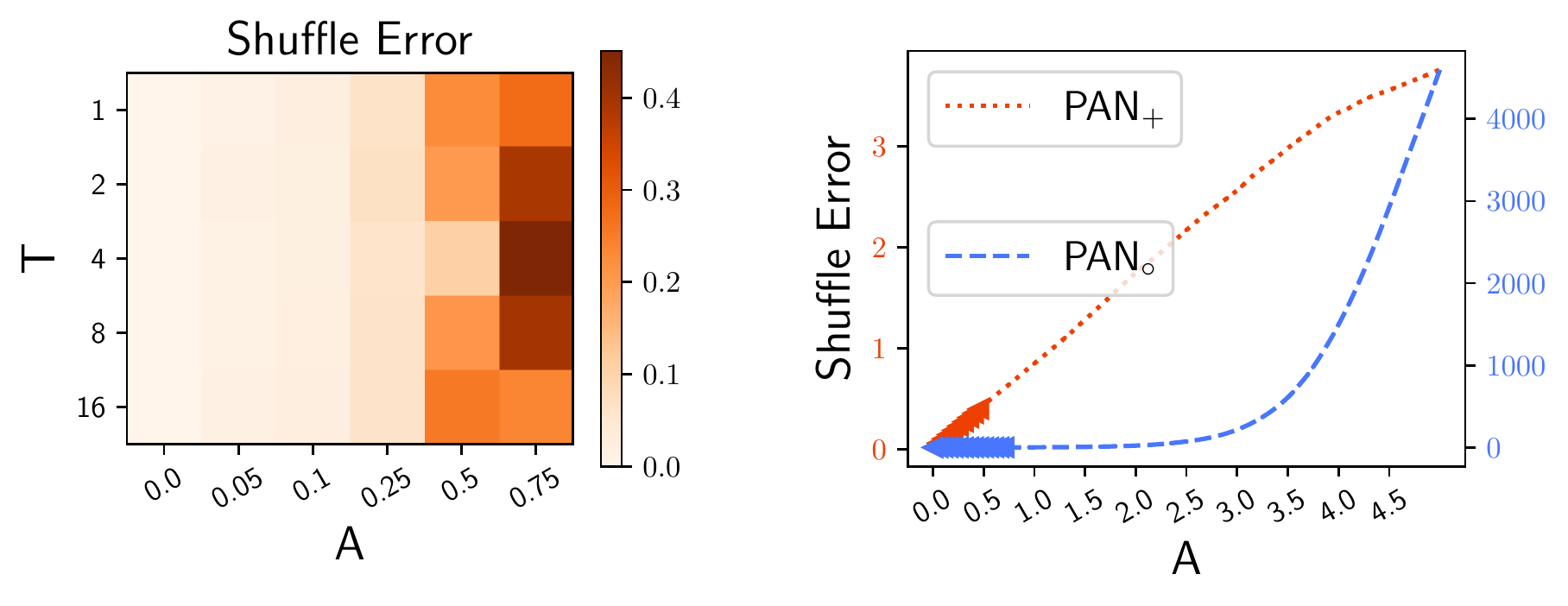}
		\caption{{\bf Left}: shuffle error (Eq.~\ref{eq:sf-error}) with various $T$ and $A$ (PAN$_\circ$). {\bf Right}: the difference between PAN$_+$ and PAN$_\circ$ ($T$=1). ({\scriptsize VGG13 is used, more networks are in Supp.})}
		\label{fig:sf-func}
	\end{figure}

	\paragraph{Shuffle Error with Random Data:} With different hyper-parameters of $T$ and $A$ in Eq.~\ref{eq:add-pe}/Eq.~\ref{eq:mul-pe}, we use random data generated from Gaussian distributions (i.e., $x_{i,\cdot} \sim \mathcal{N}(0,1)$) to calculate the shuffle error. The results based on VGG13 are shown in Fig.~\ref{fig:sf-func}. The error is more related to $A$ while less sensitive to $T$. This is intuitive because $T$ controls local volatility while neuron permutation could happen globally, e.g., the first neuron could swap positions with the last neuron. A larger $A$ leads to a larger shuffle error, i.e., the more serious the network loses the permutation invariance property. In addition, the shuffle error based on the additive PANs increases linearly, while that based on the multiplicative PANs increases quickly. This verifies the theoretical analysis in Sect.~\ref{sec:pan}. However, in practice, a larger $A$ may cause training failure and we only set $A \in [0.0, 0.25]$ for additive PANs and $A \in [0.0, 0.75]$ for multiplicative PANs (the bold part on the right side of Fig.~\ref{fig:sf-func}).
	
	\begin{figure}[t]
		\centering
		\includegraphics[width=\linewidth]{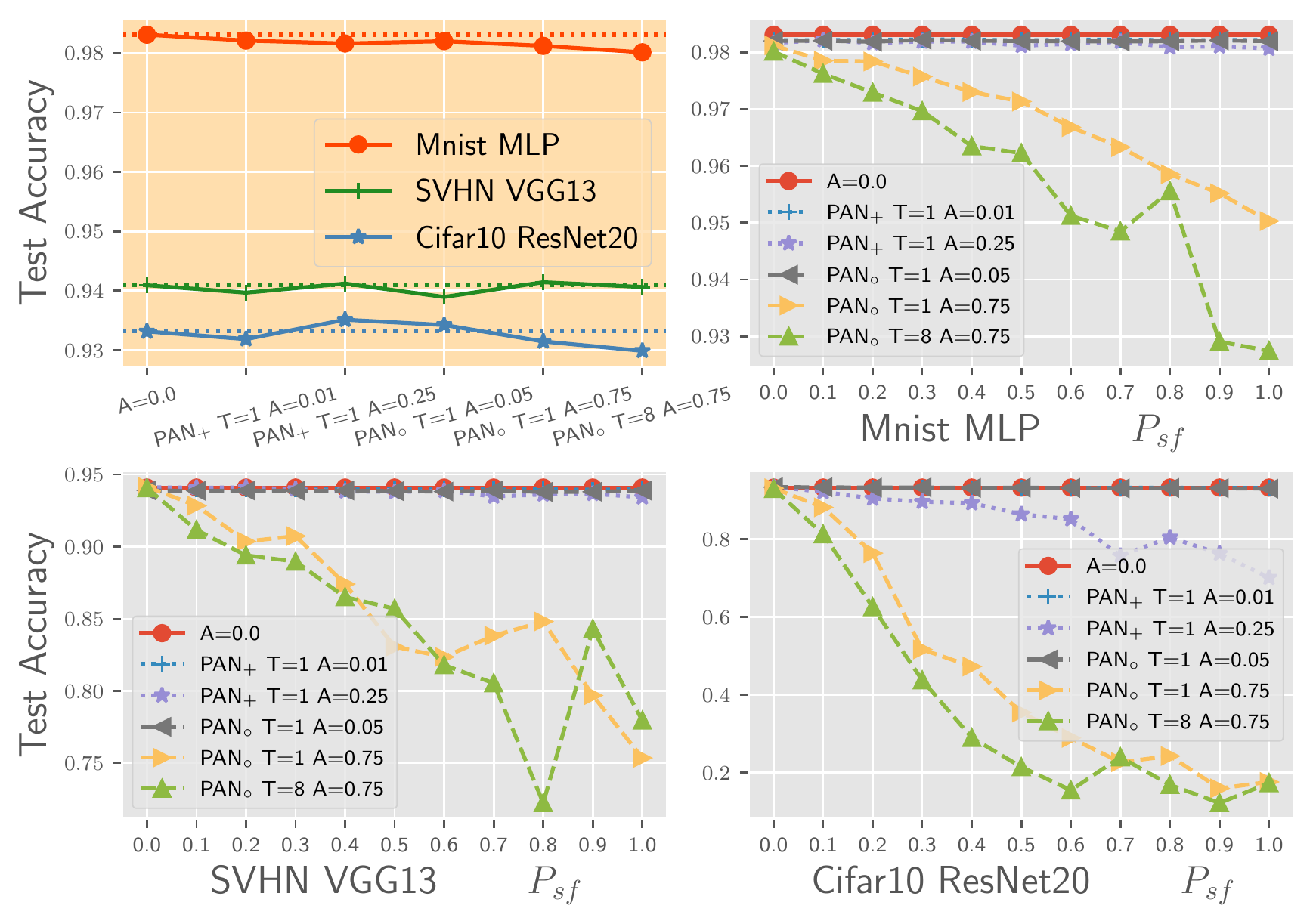}
		\caption{{\bf The first}: test accuracy of models trained with different PANs. {\bf The other three}: test accuracy change after manual permutation with various $P_{\text{sf}}$. }
		\label{fig:sf-test}
	\end{figure}

	\paragraph{Influence on Inference:} We study the influence of PANs on test accuracies. We use MLP on Mnist, VGG13 on SVHN, and ResNet20 on Cifar10. We first train models with various PANs until convergence, and the model performances are shown in the first figure of Fig.~\ref{fig:sf-test}. The horizontal dotted lines show the accuracies of normal networks, and the solid segments show the results of networks with various PANs. We find that introducing PANs to neural networks does not improve performances, but brings a slight degradation. That is, PANs could make the network somewhat harder to train. More studies of how PANs influence the network predictions could be found in Supp. Then, we investigate the shuffle error reflected by the change of test accuracies. Specifically, we shuffle the trained network to make predictions on the test set. We vary several groups of $T$ and $A$ for PANs. We show the results in the last three figures of Fig.~\ref{fig:sf-test}. With larger $P_{\text{sf}}$, i.e., more neurons are shuffled, the test accuracy of the network with $A=0.0$ does not change (the permutation invariance property). However, larger $A$ leads to more significant performance degradation ($A=0.25$ vs. $A=0.01$ for PAN$_+$; $A=0.75$ vs. $A=0.05$ for PAN$_\circ$). PAN$_\circ$ makes the network more sensitive to shuffling than PAN$_+$ (curves with ``$\circ$" degrades significantly). With different $T \in \{1,8\}$, the performance degradation is nearly the same, again showing that PANs are robust to $T$. These verify the conclusions in Sect.~\ref{sec:pan}. {\it Overall, PANs work as a tradeoff between model performances and control of permutation invariance.} 
	
	
	\begin{figure}
		\centering
		\begin{subfigure}{\linewidth}
			\includegraphics[width=\linewidth]{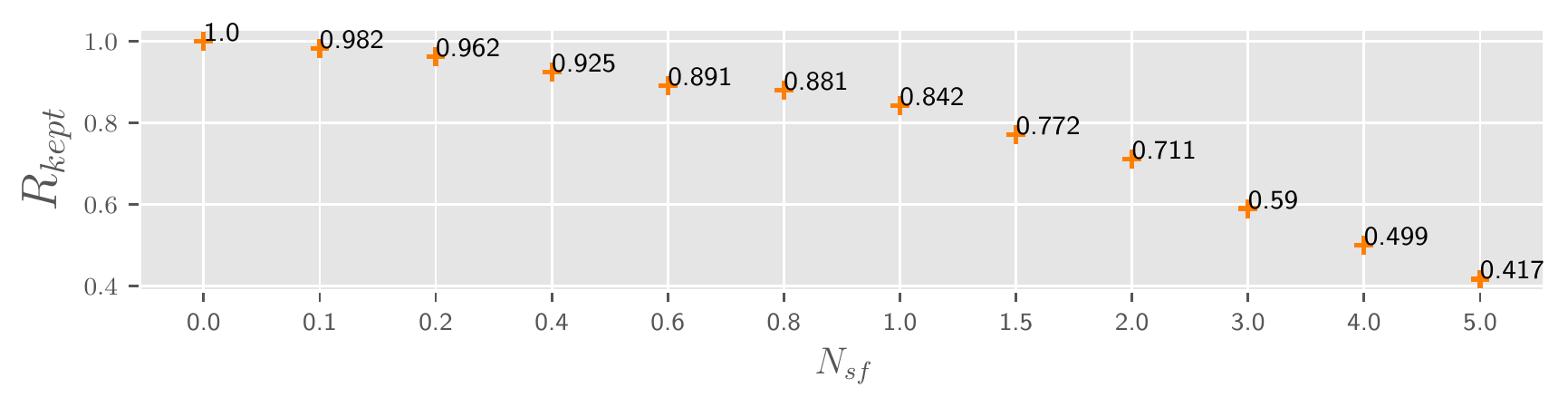}
		\end{subfigure}
		\vfill
		\begin{subfigure}{\linewidth}
			\includegraphics[width=\linewidth]{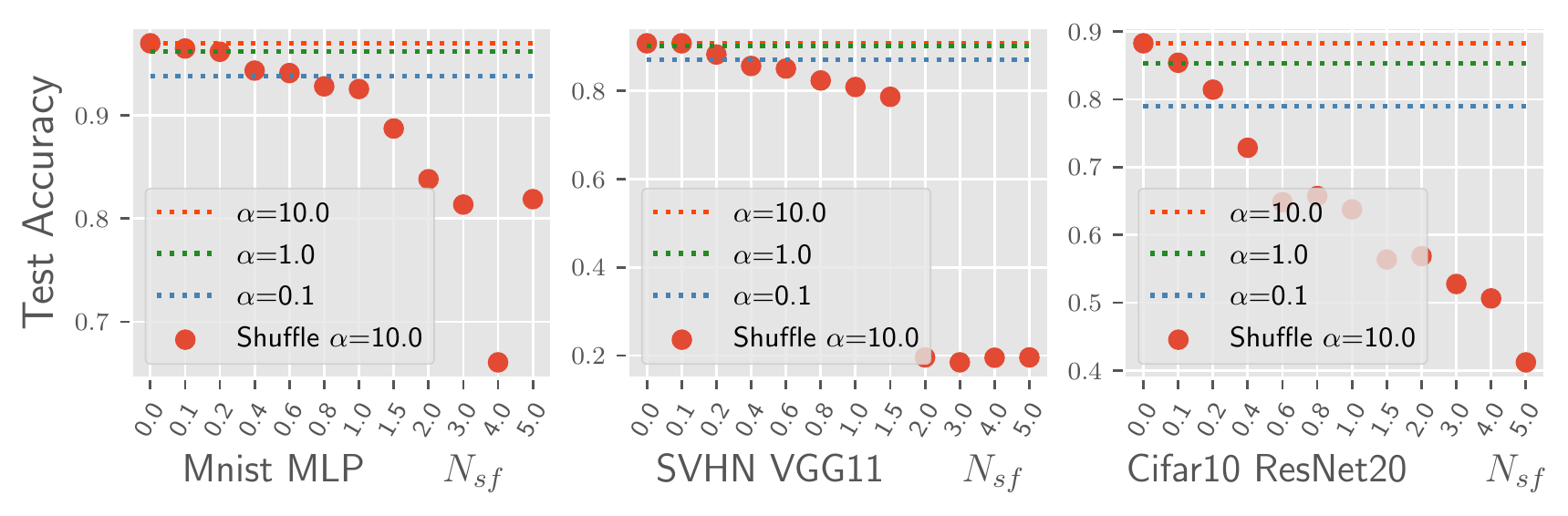}
		\end{subfigure}
		\caption{{\bf Top}: how much neurons are not shuffled with various $N_{\text{sf}}$. {\bf Bottom}: test accuracies of FL with various $\alpha$ (dotted lines) and accuracies after manual shufflling on i.i.d. data ($\alpha=10.0$) (red scatters).}
		\label{fig:sf-fl}
	\end{figure}

	\subsection{Decentralized Training}
	Then we study the effects of introducing PANs to FL. We first present some empirical studies to verify the pre-alignment effects of PANs, and then show performances.
	
	\paragraph{How many neurons are misaligned in FL?} Although some previous works~\cite{BFNM,FedMA,Fed2} declare that neurons could be dislocated when faced with non-i.i.d. data, they do not show this in evidence and do not show the degree of misalignment. We present a heuristic method: {\it we manually shuffle the neurons during local training with i.i.d. data and study how much misalignment could cause the performance to drop to the same as training with non-i.i.d. data}. Specifically, during each client's training step (each batch as a step), we shuffle the neurons with a probability $\frac{N_{\text{sf}}}{E\times N_k/B}$, where $B$,$E$,$N_k$ are respectively the batch size, the number of local epochs, and the number of local data samples. In each shuffle process, we keep $P_{\text{sf}}=0.1$. $N_{\text{sf}}$ determines how many times the network could be shuffled during local training. Larger $N_{\text{sf}}$ means more neurons are shuffled upon finishing training, e.g., $N_{\text{sf}}=1.0$ keeps approximately $84\%$ neurons not shuffled as shown in Fig.~\ref{fig:sf-fl}. The calculation of $R_{\text{kept}}$ in Fig.~\ref{fig:sf-fl} is presented in Supp. Then, we show the test accuracies of FedAvg~\cite{FedAvg} under various levels of non-i.i.d. data, i.e., $\alpha \in \{10.0, 1.0, 0.1\}$. The results correspond to the three horizontal lines in the bottom three figures of Fig.~\ref{fig:sf-fl}. The scatters in red show the performances of shuffling neurons with various $N_{\text{sf}}$. Obviously, even with i.i.d. data, the larger the $N_{\text{sf}}$, the worse the performance. This implies that neuron misalignment could actually lead to performance degradation. Compared with non-i.i.d. performances, taking Cifar10 as an example, setting $N_{\text{sf}}=0.2$ could make the i.i.d. ($\alpha$=10.0) performance degrade to the same as non-i.i.d. ($\alpha$=0.1), that is, approximately $3.8\%$ neurons are misaligned on each client. This may provide some enlightenment for the quantitative measure of how many neurons are misaligned in FL with non-i.i.d. data.
	
	\begin{figure}
		\centering
		\includegraphics[width=\linewidth]{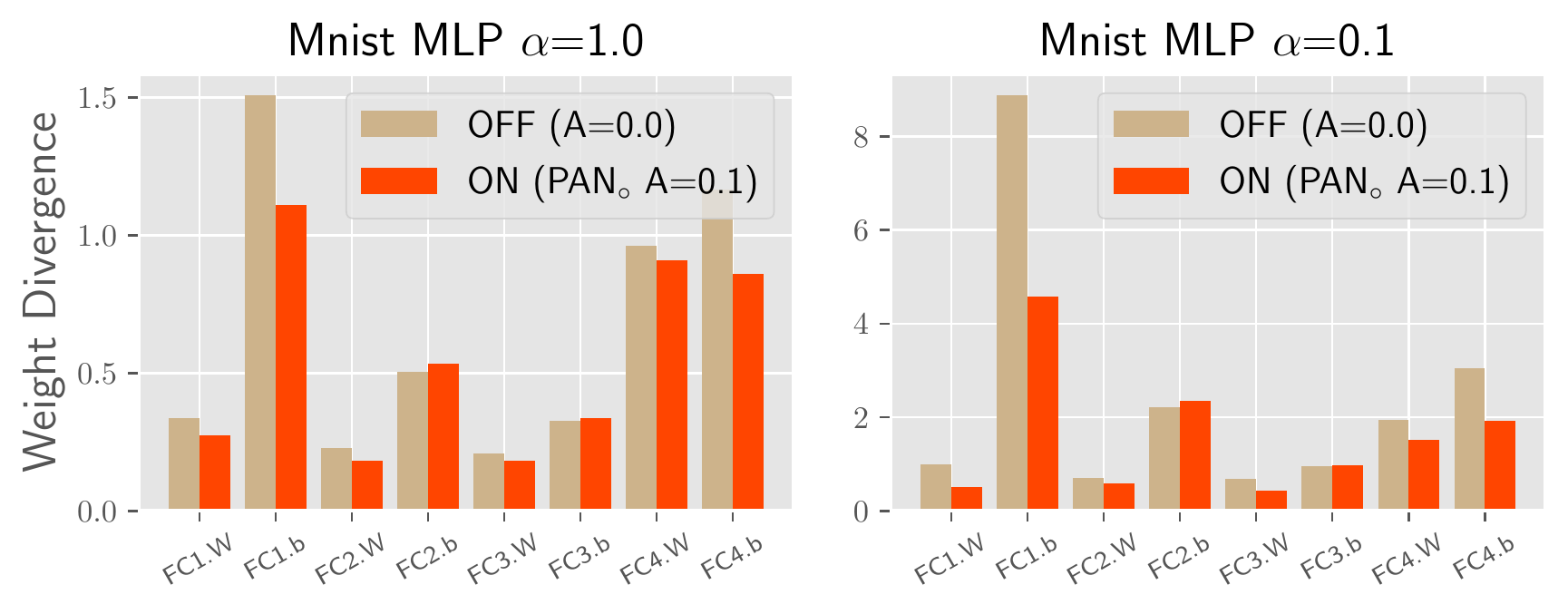}
		\caption{Weight divergence with PANs off/on. ({\scriptsize $E=5$, MLP on Mnist, more datasets' results  are in Supp.})}
		\label{fig:wdiv}
	\end{figure}

	\begin{figure}
		\centering
		\includegraphics[width=\linewidth]{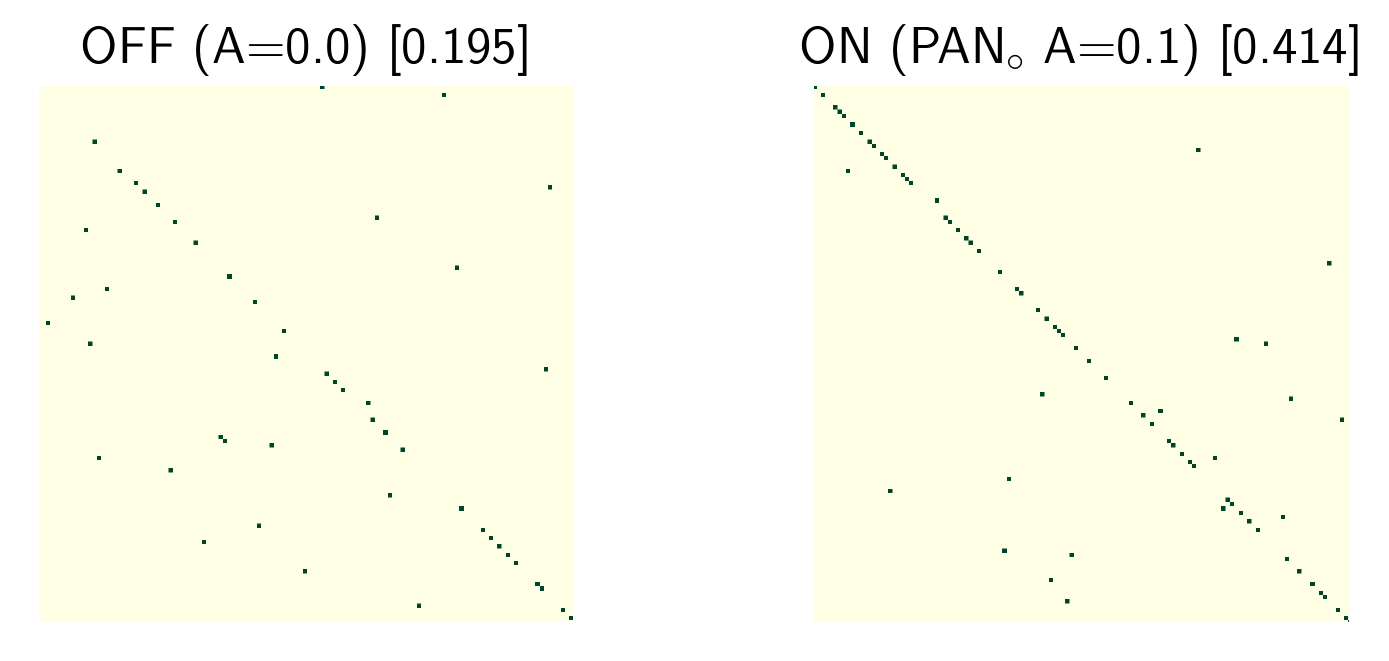}
		\caption{Optimal assignment matrix with PANs off/on, left vs. right. ({\scriptsize $\alpha=1.0$, $E=20$, VGG9 Conv5 on Cifar10, more results are in Supp.})}
		\label{fig:assign}
	\end{figure}

	\paragraph{Do PANs indeed reduce the possibility of neuron misalignment?} We propose several strategies from aspects of parameters, activations, and preference vectors to compare the neuron correspondences in FL with PANs off/on. For PANs turned on, we use multiplicative PANs with $T=1.0$ and $A=0.1$ by default.
	
	\noindent \textbf{\uppercase\expandafter{\romannumeral1}. Weight Divergence:} Weight divergence~\cite{Fed-NonIID-Data} measures the variances of local parameters. Specifically, we calculate $\frac{1}{|S_t|}\sum_{k \in S_t}\lVert W_l^{(k)} - W_l \rVert_2$ for each layer $l$. $W_l=\frac{1}{|S_t|}\sum_{k\in S_t} W_{l}^{(k)}$ denotes the averaged parameters. The weight divergences of MLP on Mnist with $\alpha \in \{1.0,0.1\}$ are in Fig.~\ref{fig:wdiv}, where PANs could reduce the divergences a lot (the red bars). This corresponds to the explanation in Sect.~\ref{sec:pan-in-fl} that clients' parameters are partially updated towards the same direction.
	
	\begin{figure}
		\centering
		\includegraphics[width=\linewidth]{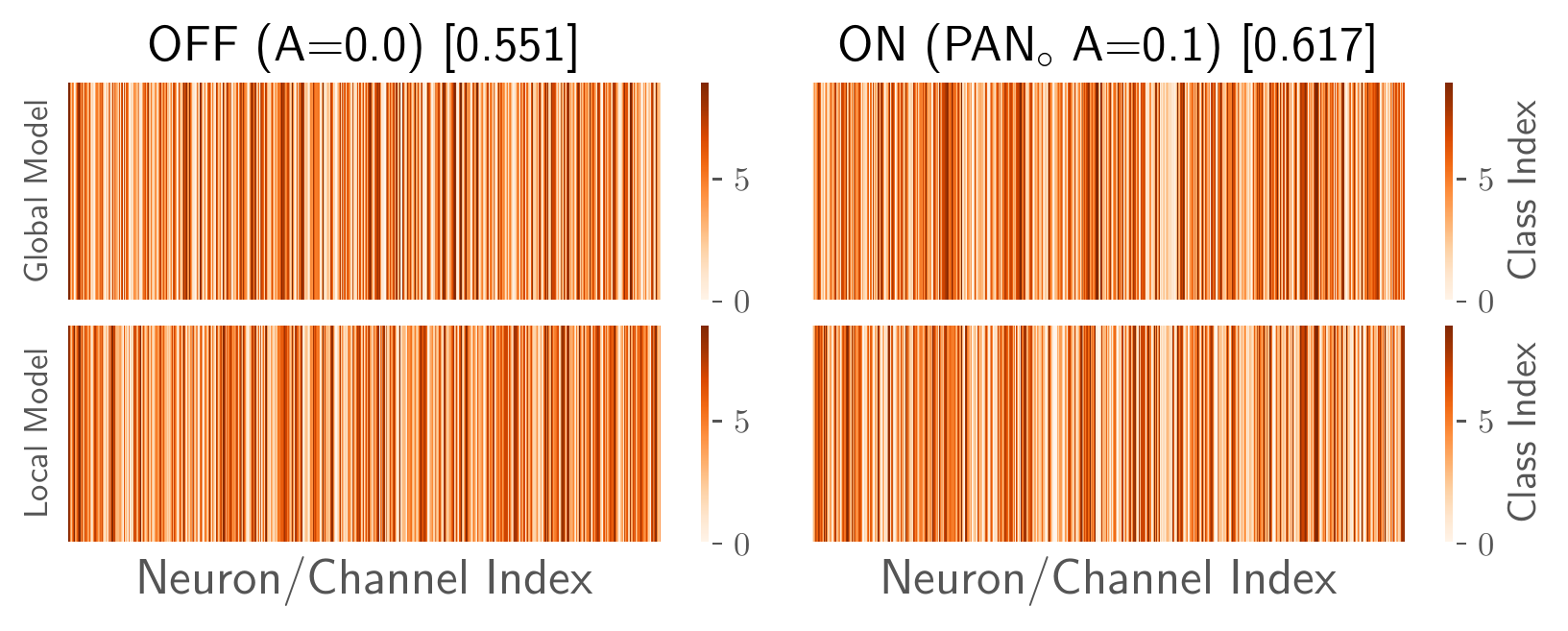}
		\caption{Preference vectors with PANs off/on, left vs. right. ({\scriptsize $\alpha=1.0$, VGG9 Conv6 on Cifar10, more results are shown in Supp.})}
		\label{fig:prefer}
	\end{figure}
	
	\noindent \textbf{\uppercase\expandafter{\romannumeral2}. Matching via Optimal Assignment:} We feed 500 test samples into the network and obtain the activations of each neuron as its representation. Neurons' representations of global and local model are denoted as $h_{l} \in \mathcal{R}^{J_l \times m}$ and $h_{l}^{(k)} \in \mathcal{R}^{J_l \times m}$, where $m=500$. Then we search for the optimal assignment matrix $Q \in \{0,1\}^{J_l \times J_l}$ that minimizes $\sum_{i=1}^{J_l} \sum_{j=1}^{J_l} Q_{ij} \lVert h_{l,i} - h_{l,j}^{(k)} \rVert_2$ and satisfies $\sum_{i}Q_{i,\cdot}=1$, $\sum_{j} Q_{\cdot,j}=1$. In fact, $Q$ is a permutation matrix that could approximately reflect the disturbance of neurons, and it could match neurons with similar outputs. We plot the solved matching matrix in Fig.~\ref{fig:assign}, where the number in ``[]" shows the ratio of the diagonal ones. Using PANs could make the diagonal denser, implying that neurons at the same coordinates output similarly.
	
	\noindent \textbf{\uppercase\expandafter{\romannumeral3}. Visualizing Neurons via Preference Vectors:} Then, we correspond neurons to classes via calculating preference vectors as done in~\cite{Fed2}. Specifically, we calculate $p_c=\sum_{b=1}^B \text{Acti}(x_{c,b})\frac{\partial Z_c}{\partial \text{Acti}(x_{c,b})}$ for each class $c$, and then concatenate all classes as the preference vector $[p_1,p_2,\cdots,p_C]$. $\text{Acti}(\cdot)$ denotes the activation value and $Z_c$ is the prediction score of the $c$th class. Then, $\arg\max_c p_c$ implies which class the neuron contributes to more. The results are shown in Fig.~\ref{fig:prefer}, where each vertical line represents a neuron/channel. The number in ``[]" shows how much neurons/channels correspond to the same class between global and local models. With PANs, the coordinate matching results are better. These empirical results verify the pre-alignment effects brought by PANs.
	
	\begin{figure}
		\centering
		\includegraphics[width=\linewidth]{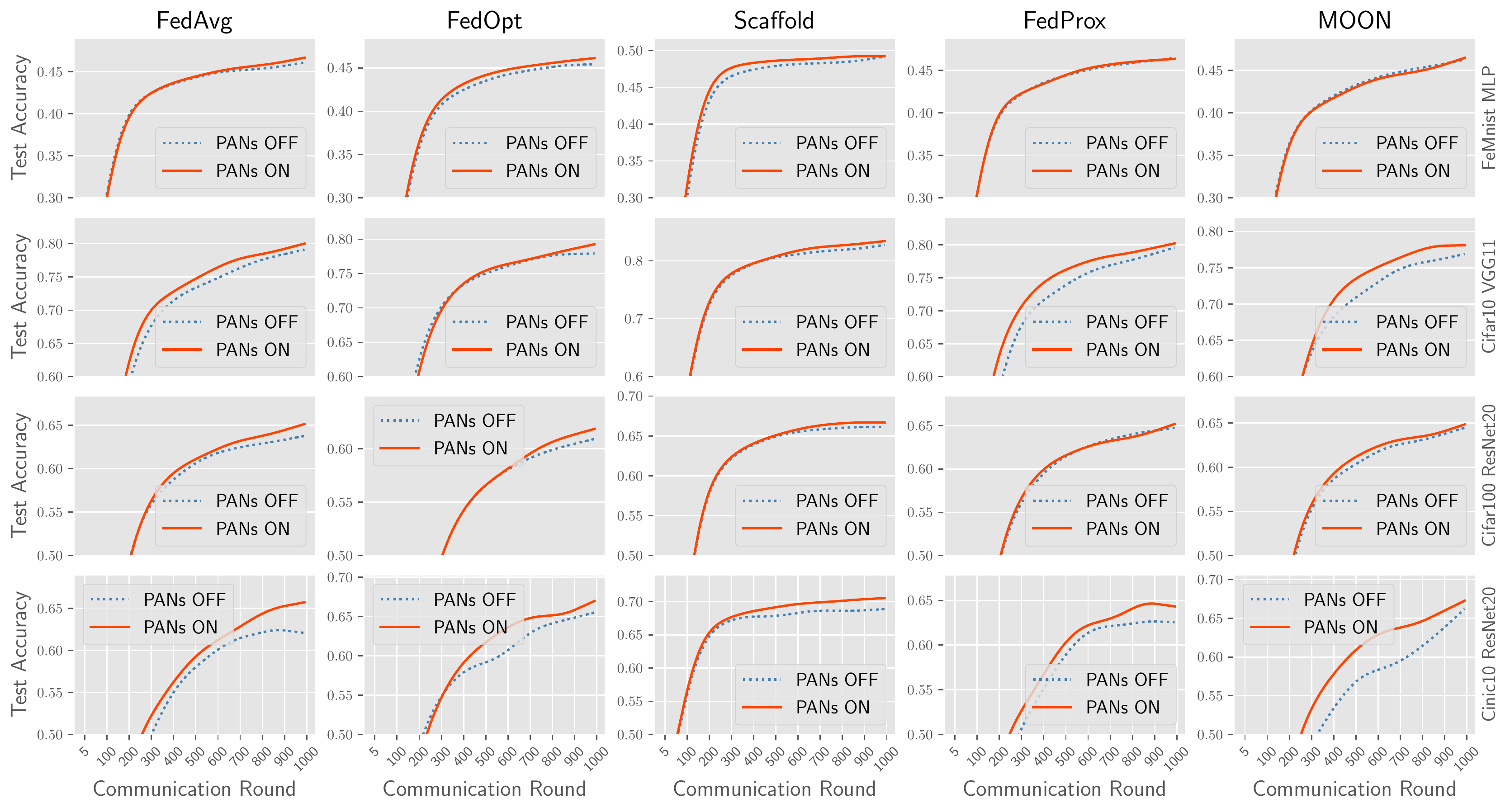}
		\caption{Comparison results on non-i.i.d. data ($\alpha$=0.1). Rows show datasets and columns show FL algorithms. PANs could universally improve these algorithms. ({\scriptsize More datasets are shown in Supp.})}
		\label{fig:compare-noniid}
	\end{figure}
	
	\begin{table*}
		\centering
		\begin{tabular}{@{}c|ccccc|cc@{}}
			\toprule
			Settings ($K,R,\alpha,E$) & FedAvg & FedProx & FedMA & Fed$^2$ & FedDF & FedAvg$^\star$ & FedAvg$^\star$+PANs  \\
			\midrule
			($16,1.0,0.5,20$) & 86.29 & 85.32 & 84.0 (\small{87.53, $E=150$}) & 88.29 & - & 86.83 & {\bf 88.49$\pm$0.07} \\
			($20,0.4,1.0,40$) & 78.34 & 78.60 & 65.0 & - & 80.36 & 79.76 & {\bf 81.94$\pm$0.09} \\
			\bottomrule
		\end{tabular}
		\caption{Comparison results with other popular FL algorithms on Cifar10 with VGG9. The left shows settings. The middle shows the cited results from FedMA~\cite{FedMA}, Fed$^2$~\cite{Fed2}, and FedDF~\cite{FedDF}. The last two columns show the results we implement.}
		\label{tab:compare-other}
	\end{table*}

	\begin{table}[tb]
		\centering
		\begin{tabular}{@{}c|cc|c@{}}
			\toprule
			 ($K,R,\alpha,E$) & FedMA$^\star$ & Fed$^{2}$$^\star$ & FedAvg$^\star$+PANs  \\
			\midrule
			($16,1.0,\underline{0.1},20$) & 83.91 & 82.26  & {\bf 85.82 $\pm$0.16} \\
			($16,\underline{0.4},0.5,20$) & 48.25 & 81.23 & {\bf 82.87 $\pm$0.21} \\
			\bottomrule
		\end{tabular}
		\caption{Comparison results with SOTA on more scenes. The results are all implemented by our reproduced code.}
		\label{tab:compare-more}
	\end{table}
	
	\paragraph{Do PANs bring performance improvement in FL?} We then compare the performances of FL with PANs off/on.
	
\begin{figure}
		\centering
		\includegraphics[width=\linewidth]{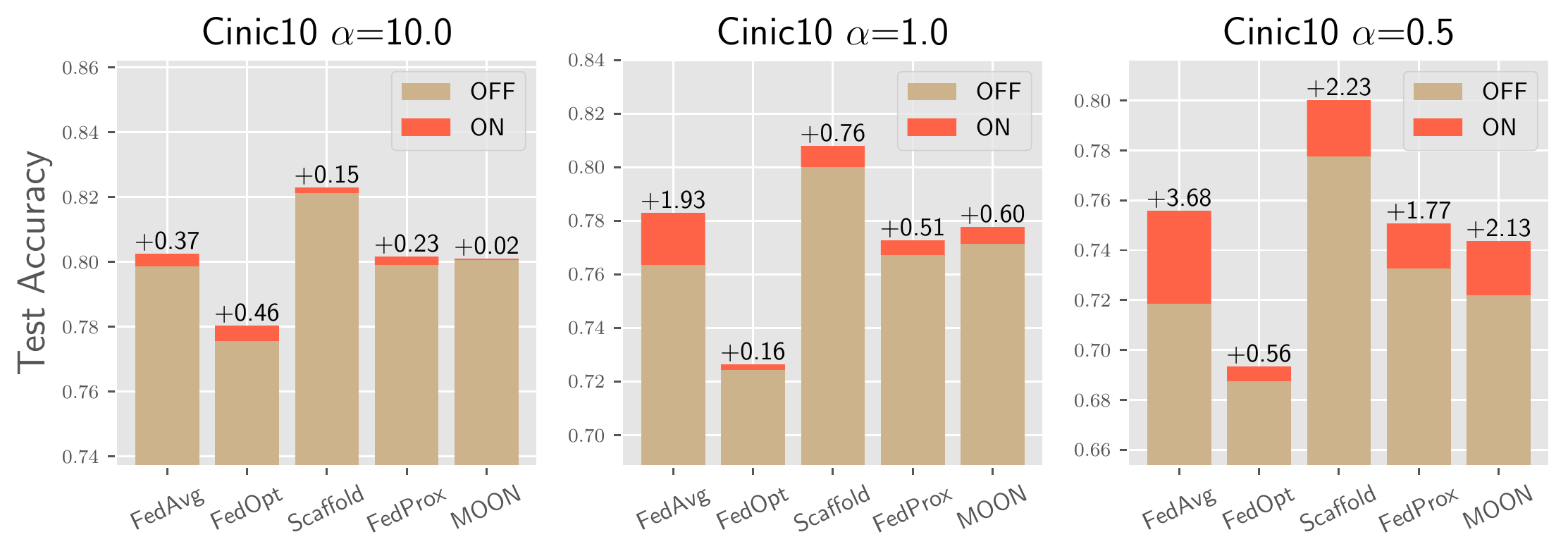}
		\caption{Comparisons under various levels of non-i.i.d. data on Cinic10. Smaller $\alpha$ implies more non-i.i.d. data. ({\scriptsize More datasets are shown in Supp.})}
		\label{fig:compare-alpha}
	\end{figure}

\begin{figure}
		\centering
		\includegraphics[width=\linewidth]{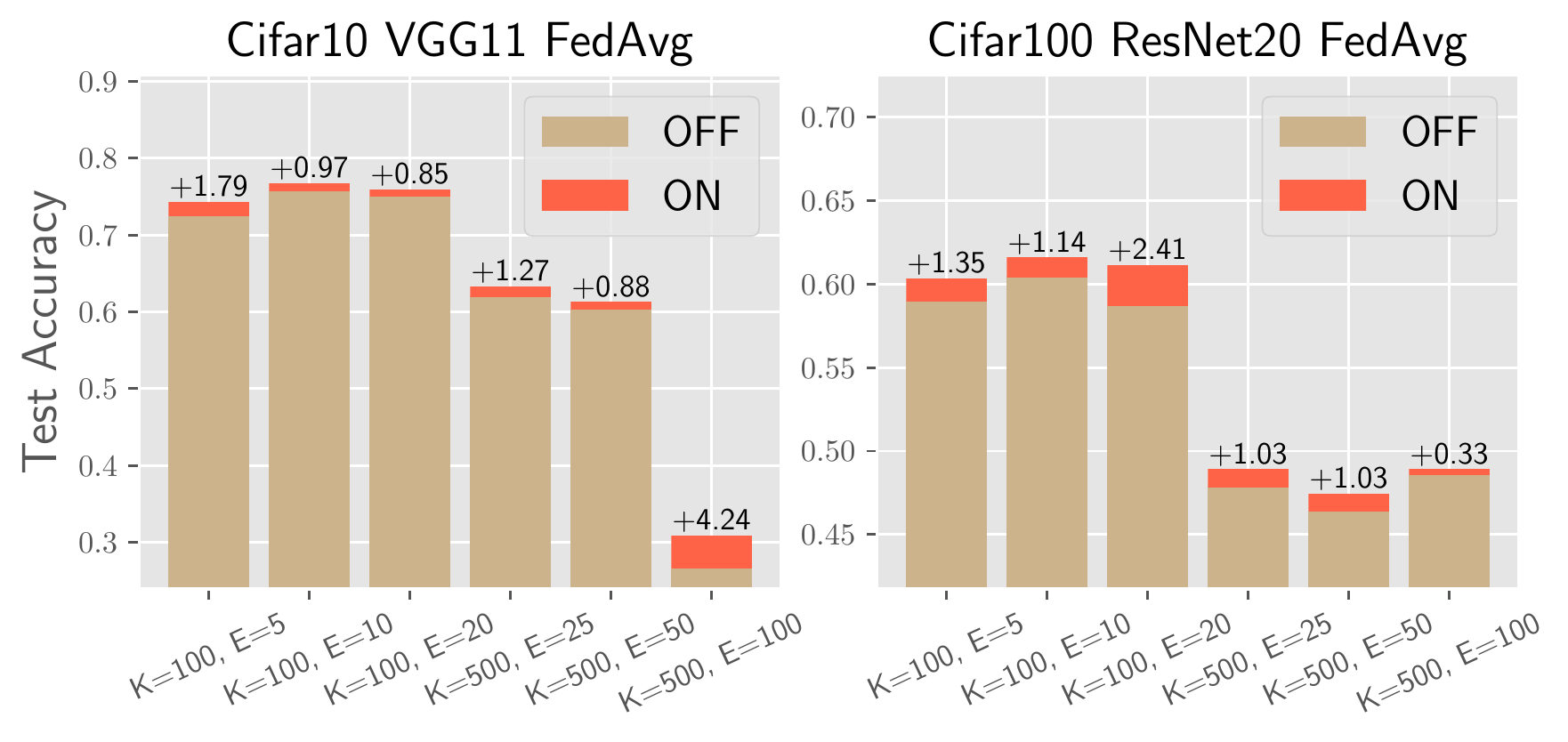}
		\caption{Comparisons under different FL scenes ($K$, $E$) based on FedAvg. ({\scriptsize Scaffold results are shown in Supp.})}
		\label{fig:compare-scenes}
	\end{figure}

	\noindent \textbf{\uppercase\expandafter{\romannumeral1}. Universal Application of PANs:} We first apply PANs to some popular FL algorithms as introduced in Sect.~\ref{sec:relate}, including FedAvg~\cite{FedAvg}, FedProx~\cite{FedProx}, FedOpt~\cite{FedOpt}, Scaffold~\cite{Scaffold}, MOON~\cite{MOON}. These methods solve the non-i.i.d. problem from different aspects. Training details of these algorithms are provided in Supp. We add PANs to them and investigate the performance improvements on FeMnist, Cifar10, Cifar100, and Cinic10, where $\alpha=0.1$, $K=100$, $R=10\%$, $E=5$, $H=1000$. We use $A=0.0$ as the baseline. Hyper-parameters are searched from three groups: PAN$_+$ with $A=0.05$, PAN$_\circ$ with $A=0.05$, PAN$_\circ$ with $A=0.1$, and the best result is reported in Fig.~\ref{fig:compare-noniid}. PANs indeed improve these algorithms. With various non-i.i.d. levels of decentralized data, i.e., $\alpha \in \{10.0, 1.0, 0.5\}$. We report the averaged accuracy of the last five communication rounds in Fig.~\ref{fig:compare-alpha} ($H=200$ communication rounds with other hyper-parameters the same). Obviously, more non-i.i.d. scenes (smaller $\alpha$) experience more significant improvements. This is related to the regularization effect as analyzed in Sect.~\ref{sec:pan-in-fl}. We also investigate the results with various numbers of clients and local training epochs, i.e., $K$ and $E$. The results of FedAvg on Cifar10 and Cifar100 are shown in Fig.~\ref{fig:compare-scenes}, where we take $\alpha=0.1$ and $H=400$. On average, introducing PANs could lead to about $1\%$ to $2\%$ improvement on various scenes. These studies verify that PANs could be universally and effectively applied to FL algorithms under various settings.
	
\begin{figure}
		\centering
		\begin{subfigure}{0.48\linewidth}
			\includegraphics[width=\linewidth]{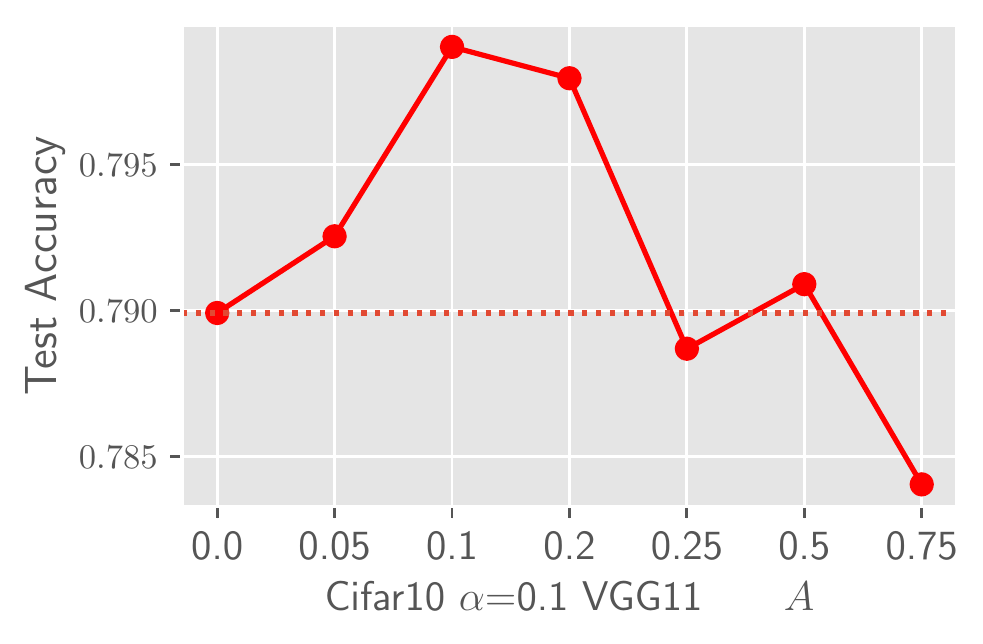}
		\end{subfigure}
		\hfill
		\begin{subfigure}{0.48\linewidth}
			\includegraphics[width=\linewidth]{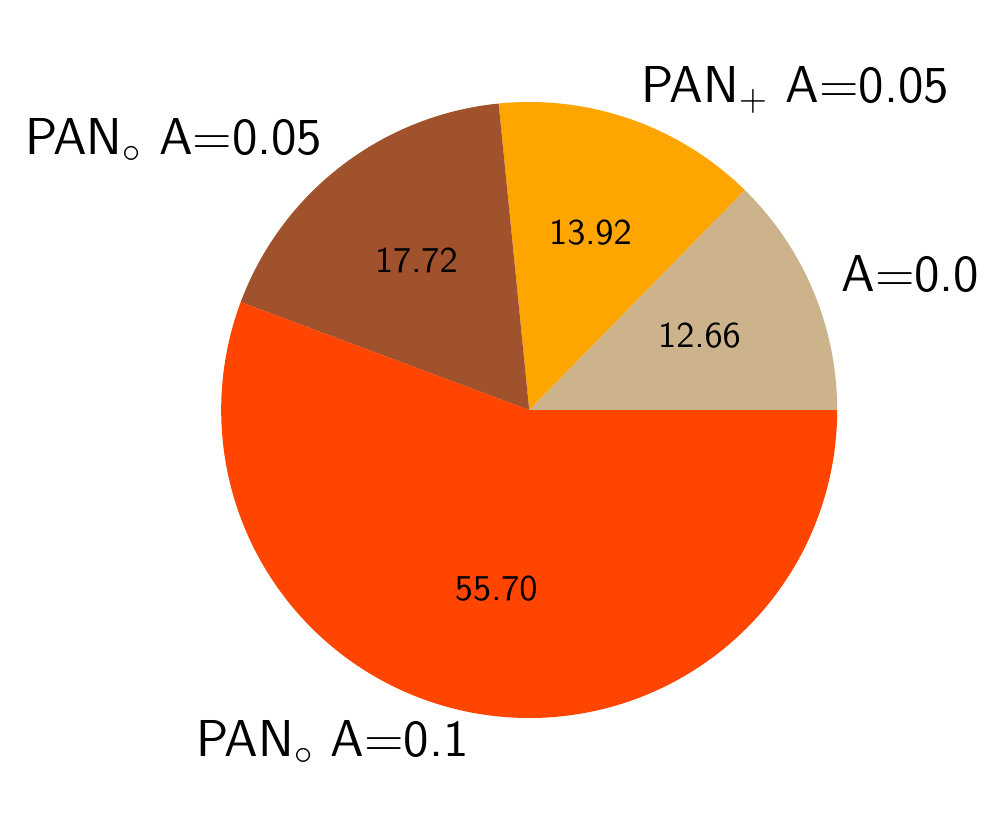}
		\end{subfigure}	
		\caption{{\bf Left}: performance comparisons under various $A$. {\bf Right}: the distributions of optimal hyper-parameters.}
		\label{fig:compare-hyper}
	\end{figure}

	\noindent \textbf{\uppercase\expandafter{\romannumeral2}. Hyper-parameter Analysis:} We first vary $A$ on Cifar10 and plot the results on the left of Fig.~\ref{fig:compare-hyper}. We set $T=1.0$ and only report the results of multiplicative PANs. Setting $A$ around 0.1 could improve the performance a lot, while using larger $A$ experiences degradation, which is because neural networks become harder to train. This again shows that $A$ is a tradeoff between neuron pre-alignment and network performance. The proportions of the optimal hyper-parameters from the results of the above experiments are shown on the right of Fig.~\ref{fig:compare-hyper}. Using $A=0.1$ in multiplicative PANs is a good choice. $A=0.0$ means turning off PANs, and its ratio is only about $13\%$, which means turning on PANs is useful in most cases.
	
	
	\noindent \textbf{\uppercase\expandafter{\romannumeral3}. Comparing with SOTA:} FedMA~\cite{FedMA} and Fed$^2$~\cite{Fed2} are representative works that solve the parameter alignment problems in FL. We collect the reported settings and results in FedMA, Fed$^2$, and FedDF~\cite{FedDF}, and compare the performances under the same settings. We list the results on Cifar10 with VGG9 in Tab.~\ref{tab:compare-other}, where the last three columns show our results. Although our reproduced FedAvg performs slightly better than the cited results, the performance gain via introducing PANs is remarkable. We then vary the settings of ($16,1.0,0.5,20$) from two aspects: (1) decreasing the non-i.i.d. $\alpha$ from $0.5$ to $0.1$, i.e., a more non-i.i.d. scene; (2) decreasing the client selection ratio from $1.0$ to $0.4$, i.e., partial client participation. Aside from the above changes, other hyper-parameters are kept the same. We run the code provided by FedMA\footnote{\url{https://github.com/IBM/FedMA}} and reproduce Fed$^2$ via our implementations. The results are listed in Tab.~\ref{tab:compare-more}. FedMA performs especially worse under partial client participation. Fed$^2$ also performs not so well. Our method surpasses the compared methods obviously in these cases. Furthermore, our method is more efficient, e.g., with four 10-core Intel(R) Xeon(R) Silver 4210R CPUs @ 2.40GHz and one NVIDIA GeForce RTX 3090 GPU card, FedMA needs about 4 hours for a single communication round while ours only requires several minutes.
	
	\noindent \textbf{\uppercase\expandafter{\romannumeral4}. More Studies:} We study using optimal transport to fuse neural networks with PANs as done in~\cite{OTFusion}. We also investigate the BatchNorm~\cite{BN} and GroupNorm~\cite{GN} used in VGG or ResNet, where PANs are more applicable to BatchNorm. We finally investigate some varieties of PANs for better personalization in FL~\cite{PersonalizeMAML}. These are provided in Supp.
	
	\noindent \textbf{\uppercase\expandafter{\romannumeral5}. Disadvantages:} Fusing different values makes the magnitudes of neuron activations/gradients varied, which requires a customized neuron-aware optimizer. In supp, we try applying the adaptive optimizer Adam~\cite{Adam} to PANs, but we do not find too much improvement. Hence, advanced optimizers should be explored in future work.
	
	
	\section{Conclusions}
	We propose position-aware neurons (PANs) to disable/enable the permutation invariance property of neural networks. PANs bind themselves in their positions, making parameters pre-aligned in FL even faced with non-i.i.d. data and facilitating the coordinate-based parameter averaging. PANs keep the same position encodings across clients, making local training contains consistent ingredients. Abundant experimental studies verify the role of PANs in parameter alignment. Future works are to find an optimization method specifically suitable for PANs, and extend PANs to large-scale FL benchmarks or more scenarios that require parameter alignment.
	
	\section*{Acknowledgements}
	This work is partially supported by National Natural Science Foundation of China (Grant No. 41901270), NSFC-NRF Joint Research Project under Grant 61861146001, and Natural Science Foundation of Jiangsu Province (Grant No. BK20190296). Thanks to Huawei Noah’s Ark Lab NetMIND Research Team and CAAI-Huawei MindSpore Open Fund (CAAIXSJLJJ-2021-014B). Thanks for Professor Yang Yang's suggestions. Professor De-Chuan Zhan is the corresponding author.
	
	\newpage
	{\small
		\bibliographystyle{ieee_fullname}
		\bibliography{fedpan}
	}

	\newpage
	\appendix
	\section{Dataset Details}
	\label{sec:dataset}
	
	The utilized datasets include Mnist~\cite{mnist}, FeMnist~\cite{LEAF}, SVHN~\cite{Svhn}, GTSRB~\cite{GTSRB}, Cifar10/100~\cite{cifar}, and Cinic10~\cite{Cinic10}. We detail these datasets as follows.
	\begin{itemize}
		\item \textbf{Mnist}~\cite{mnist} is a digit recognition dataset that contains 10 digits to classify. The raw set contains 60,000 samples for training and 10,000 samples for evaluation. The image size is $28 \times 28$.
		\item \textbf{SVHN}~\cite{Svhn} is the Street View House Number dataset which contains 10 numbers to classify. The raw set contains 73,257 samples for training and 26,032 samples for evaluation. The image size is $32 \times 32$.
		\item \textbf{GTSRB}~\cite{GTSRB} is the German Traffic Recognition Benchmark with 43 traffic signs. The raw set contains 39,209 samples for training and 12,630 samples for evaluation. We resize the images to $32 \times 32$.
		\item \textbf{Cifar10} and \textbf{Cifar100}~\cite{cifar} are subsets of the Tiny Images dataset and respectively have 10/100 classes to classify. They consist of 50,000 training images and 10,000 test images. The image size is $32\times 32$.
		\item \textbf{Cinic10}~\cite{Cinic10} is a combination of Cifar10 and ImageNet~\cite{ImageNet}, which contains 10 classes. It contains 90,000 samples for training, validation, and test, respectively. We do not use the validation set. The image size is $32 \times 32$.
		\item \textbf{FeMnist}~\cite{LEAF} is built by partitioning the data in Extended MNIST~\cite{EMNIST} based on the writer of the digit/character. There are 62 digits and characters in all. The total number of training samples is 805,263. There are 3,550 users, and each user owns 226.8 samples on average. We only use $10\%$ users (i.e., 355 users). For each user, we take $20\%$ of the samples to construct the global test set. We resize the images to $28 \times 28$.
	\end{itemize}
	
	For centralized training, we correspondingly use the training set and test set for the first six datasets. For FeMnist, we centralize users' training samples as the training set. For decentralized training (i.e., FL), we split the training set of the first six datasets according to Dirichlet distributions as done in previous FL works~\cite{FedDF,NonIID-Quag,FedMA}. Specifically, we split the training set onto $K$ clients and each client's label distribution is generated from $\text{Dirichlet}(\alpha)$. While for FeMnist, we directly take the 355 users as clients. Some of these datasets are utilized in previous FL works. For example, Cifar10/Cifar100/Cinic10 are recommended by FedML~\cite{FedML}, and FeMnist is recommended by LEAF~\cite{LEAF}. 
	
	\begin{figure}[t]
		\centering
		\includegraphics[width=\linewidth]{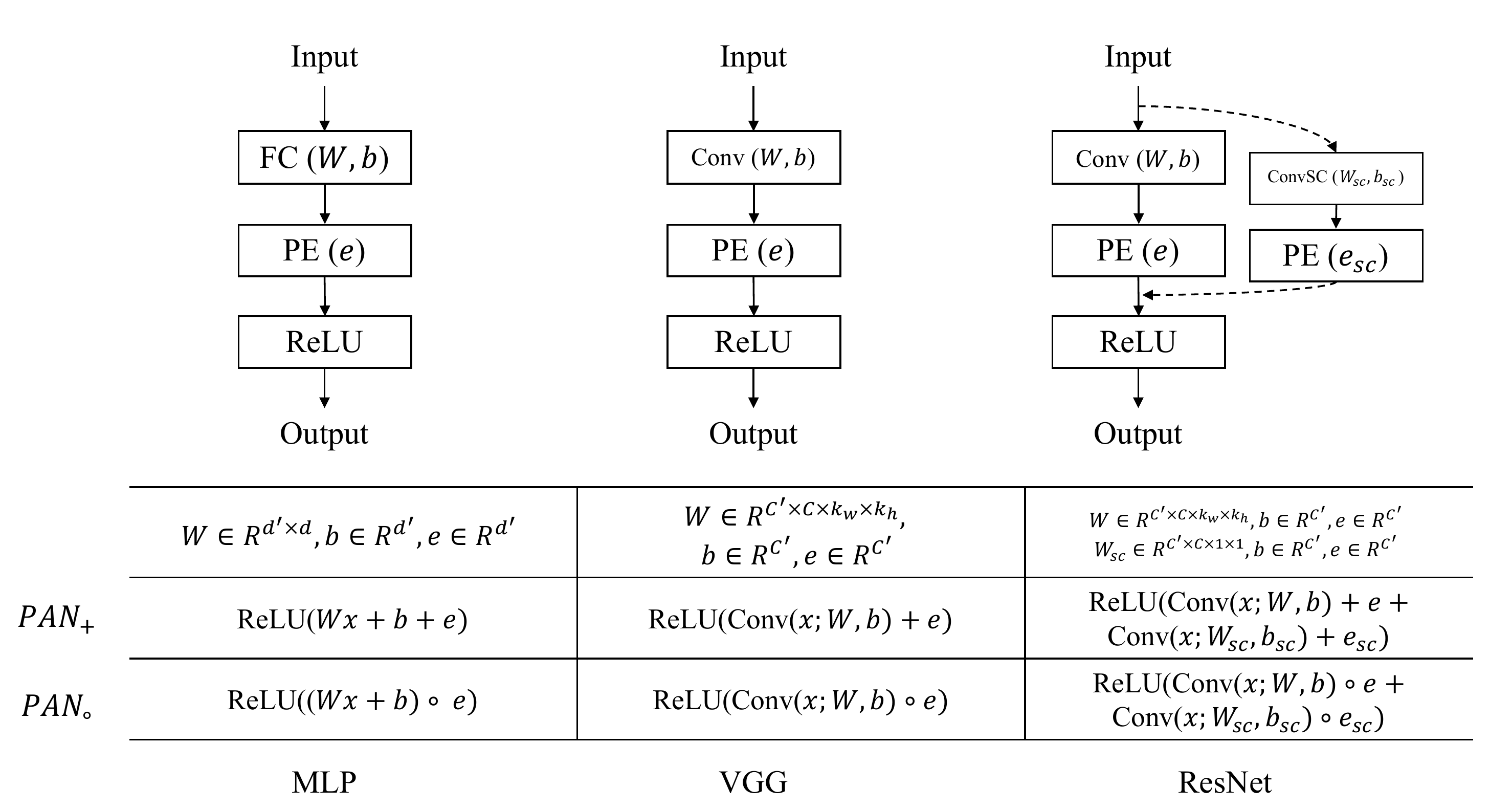}
		\caption{Network architectures with PANs. ``PE" denotes position encoding; ``SC" denotes shortcut. For ResNet, we only show one convolution layer in the basic block and omit the BatchNorm layers for simplification.}
		\label{fig:nets}
	\end{figure}

	\begin{algorithm}[tb]
		\caption{Shuffle Process}
		\label{algo:sf}
		\begin{algorithmic}[1]
			\STATE Input: parameters $\{W_{l}, b_{l}\}_{l=1}^L$; shuffle probability $P_{\text{sf}}$
			\STATE Generate-Permutation-Matrix: $\{\Pi_{l}\}_{l=1}^{L-1}$, $\Pi_{\{0,L\}}=I$
			\FOR{each layer $l = 1, 2, \ldots, L$}
			\STATE $W_l \leftarrow \Pi_l W_l \Pi_{l-1}^T$, $b_l \leftarrow \Pi_l b_l$
			\ENDFOR
		\end{algorithmic}
		
		\textbf{Generate-Permutation-Matrix}
		\begin{algorithmic}[1]
			\STATE Input: number of neurons $J$; shuffle probability $P_{\text{sf}}$
			\STATE Initialize: $\Pi=I^{J\times J}$
			\FOR{$j = 1, 2, \ldots, J$}
			\STATE sample $i$ from $\text{Range}(j + 1, J)$
			\STATE if $p \sim \text{Uniform}(0, 1) \leq P_{\text{sf}}$ then $\text{Swap}(\Pi_j, \Pi_i)$
			\ENDFOR
		\end{algorithmic}
	\end{algorithm}

	\begin{algorithm}[tb]
		\caption{Shuffle Process in FL}
		\label{algo:sf-fl}
		\begin{algorithmic}[1]
			\STATE Input: shuffle probability $P_{\text{sf}}$; expected shuffle times $N_{\text{sf}}$; number of local epochs $E$; batch size $B$; number of local samples $\{N_k\}_{k=1}^K$
			\FOR{each client $k \in S_t$}
			\STATE Calculate the number of local update steps: $r_k=E * N_k / B$
			\FOR{each local step in $[r_k]$}
			\STATE if $p \sim \text{Uniform}(0, 1) \leq N_{\text{sf}} / r_k$ run the ShuffleProcess with shuffle probability $P_{\text{sf}}$
			\ENDFOR
			\ENDFOR
		\end{algorithmic}
	\end{algorithm}

	\section{Network Details}
	\label{sec:network}
	We utilize MLP, VGG~\cite{VGG}, ResNet~\cite{ResNet} in this paper. We detail their architectures as follows:
	\begin{itemize}
		\item \textbf{MLP} denotes a multiple layer perceptron with four layers containing input and output layers. For Mnist and FeMnist, the input size is $28 \times 28 = 784$. MLP has the architecture: FC1(784, 1024), ReLU(), FC2(1024, 1024), ReLU(), FC3(1024, 1024), ReLU(), FC4(1024, $C$). $C$ denotes the number of classes.
		\item \textbf{VGG} contains a series of networks with various layers. The paper of VGG~\cite{VGG} presents VGG11, VGG13, VGG16, and VGG19. We follow their architectures and report the configuration of VGG11 as an example: 64, M, 128, M, 256, 256, M, 512, 512, M, 512, 512, M. ``M" denotes the max-pooling layer. VGG11 contains 8 convolution blocks and three fully-connected layers in~\cite{VGG}. However, we only use one fully-connected layer for classification in this paper. VGG9 is commonly utilized in previous FL works~\cite{FedMA,FedDF}, whose configuration is: 32, 64, M, 128, 128, M, 256, 256, M. We keep all the fully-connected layers in VGG9 for a fair comparison with other works. The three fully-connected layers in VGG9 are: FC(4096, 512), ReLU(), FC(512, 512), ReLU(), FC(512, $C$). We name the $i$th convolution layer in VGG as ``Conv$i$". We do not use BatchNorm~\cite{BN} in VGG by default.
		\item \textbf{ResNet} introduces residual connections to plain neural networks. We take the Cifar versions used in the paper~\cite{ResNet}, i.e., ResNet20 with the basic block. We set the initial channel as 64 (i.e., the output channel of the first convolution layer), and take nine continual basic blocks with 64, 64, 64, 128, 128, 128, 256, 256, 256 channels, respectively. We add a fully-connected layer for classification. We use BatchNorm~\cite{BN} in ResNet20 and add it before ReLU activation.
	\end{itemize}
	
	For these networks with PANs, we plot the demos in Fig.~\ref{fig:nets}. We add PE before the ReLU activation layer and after the BatchNorm layer. We show the formulations of additive PANs and multiplicative PANs in the table of Fig.~\ref{fig:nets}.
	
	\begin{figure}[t]
		\centering
		\includegraphics[width=\linewidth]{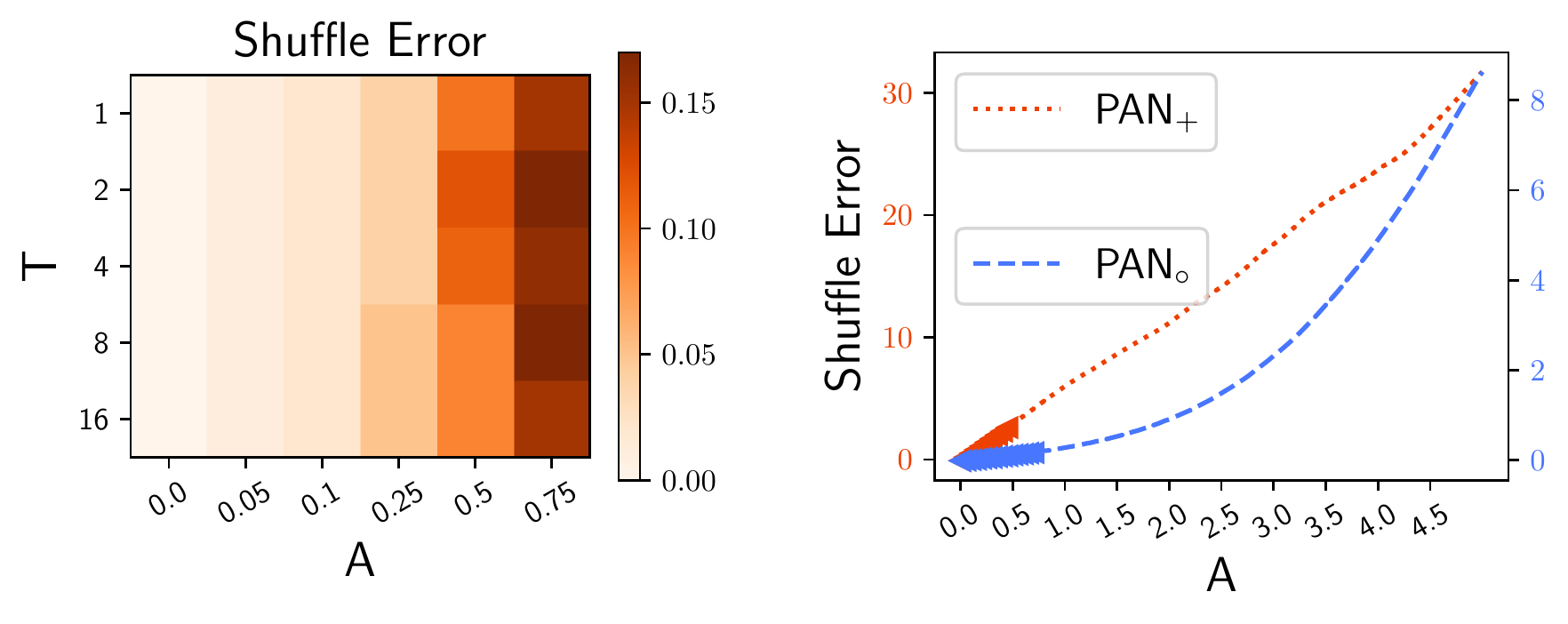}
		\caption{{\bf Left}: shuffle error with various $T$ and $A$ (PAN$_\circ$). {\bf Right}: the difference between PAN$_+$ and PAN$_\circ$ ($T$=1). ({\scriptsize MLP})}
		\label{fig:sf-func-mnist}
	\end{figure}
	
	\begin{figure}[t]
		\centering
		\includegraphics[width=\linewidth]{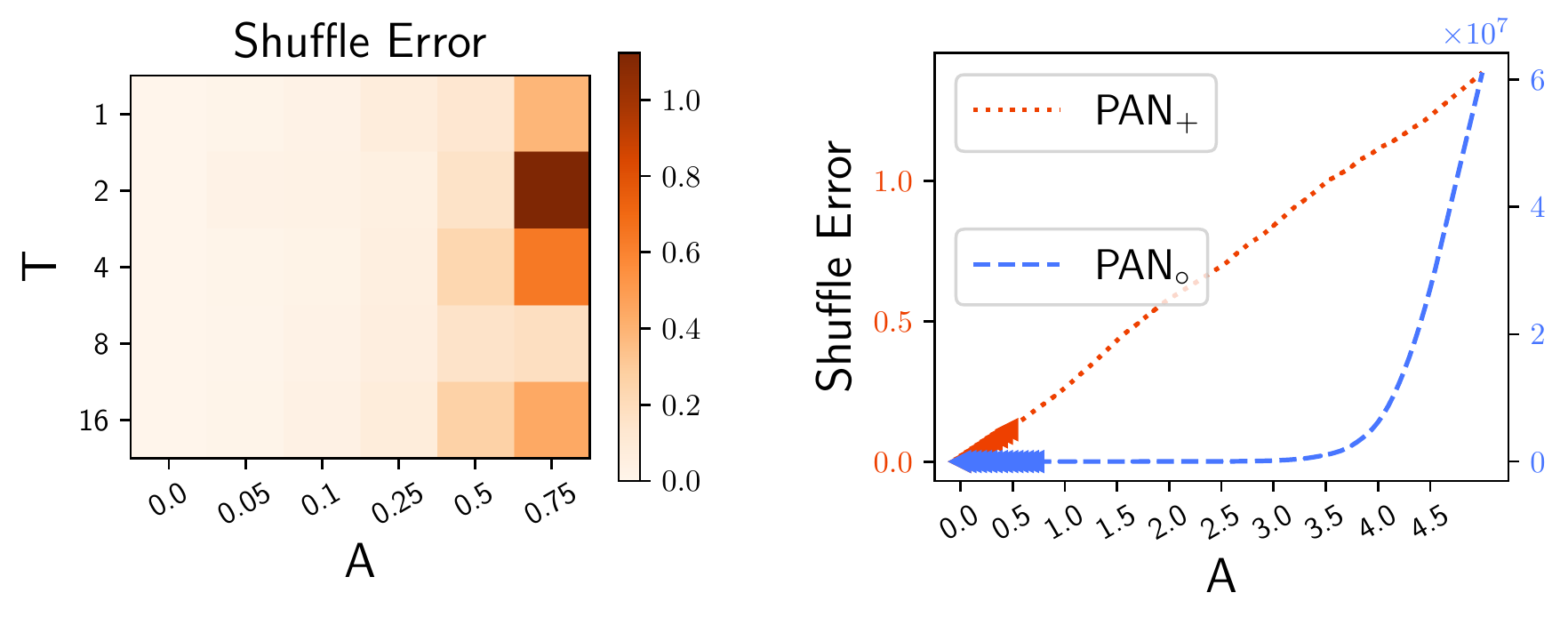}
		\caption{{\bf Left}: shuffle error with various $T$ and $A$ (PAN$_\circ$). {\bf Right}: the difference between PAN$_+$ and PAN$_\circ$ ($T$=1). ({\scriptsize ResNet20})}
		\label{fig:sf-func-cifar}
	\end{figure}
	
	\begin{figure}
		\centering
		\includegraphics[width=\linewidth]{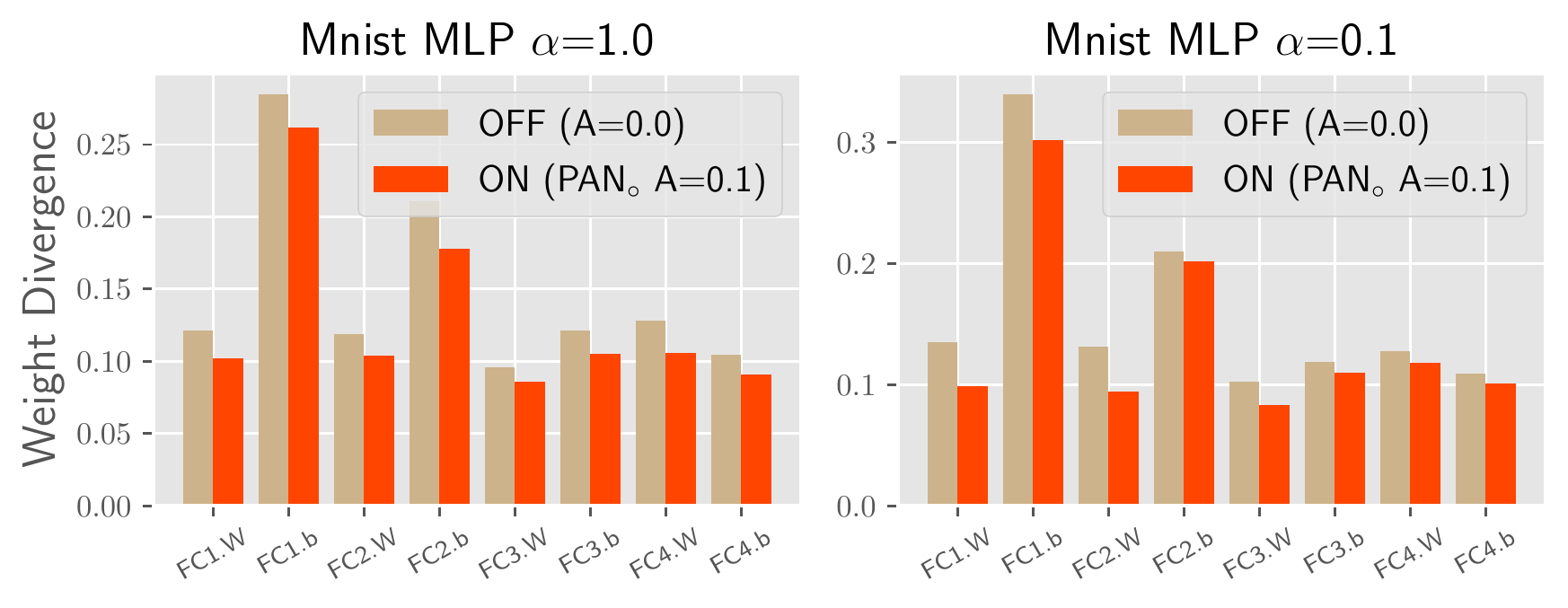}
		\caption{Weight divergence with PANs off/on. ({\scriptsize $E=20$, MLP on Mnist.})}
		\label{fig:wdiv-mnist-20}
	\end{figure}
	
	\begin{figure}
		\centering
		\includegraphics[width=\linewidth]{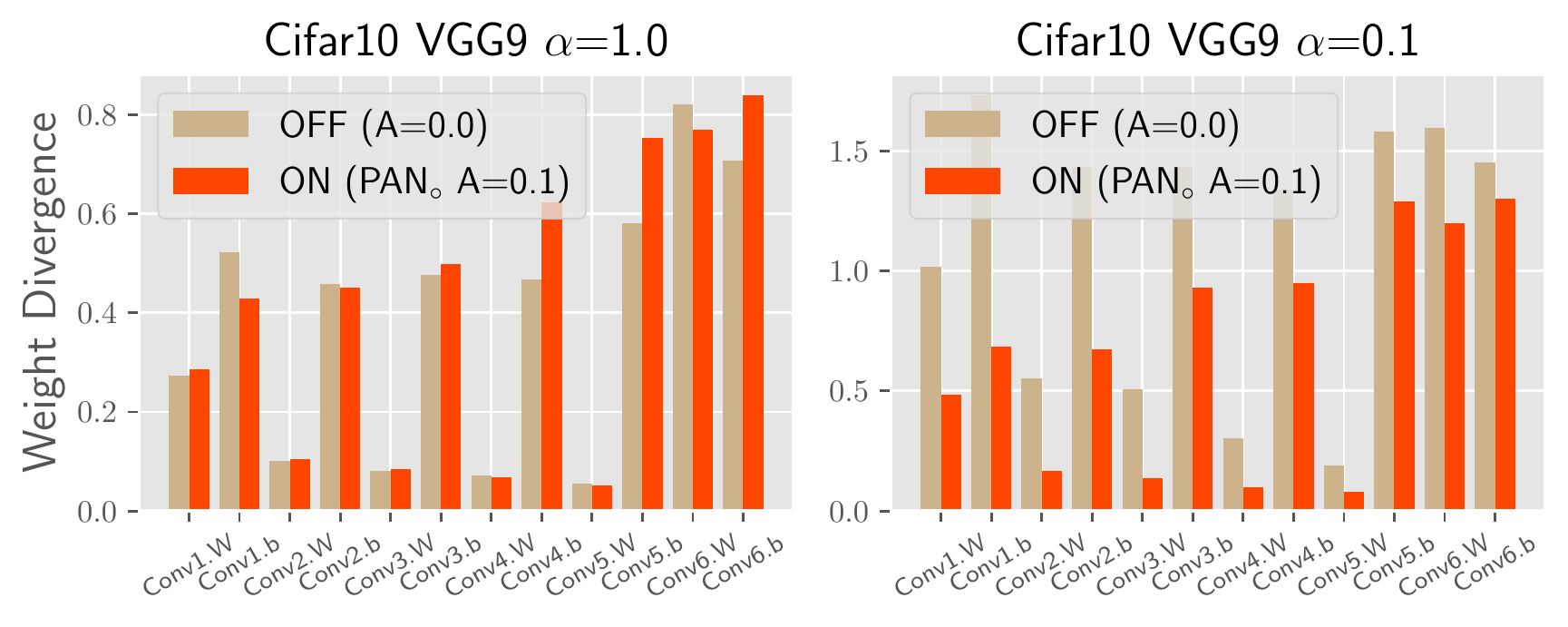}
		\caption{Weight divergence with PANs off/on. ({\scriptsize $E=5$, VGG9 on Cifar10.})}
		\label{fig:wdiv-cifar-5}
	\end{figure}

	\section{Hyper-parameter Details}
	\label{sec:training}
	For both centralized training and decentralized training (i.e., FL), we take a constant learning rate without scheduling, although some works have pointed out decaying the learning rate will help in FL~\cite{FL-Schedule}. We take SGD with momentum 0.9 as the optimizer by default if without more declaration. For MLP and VGG networks, we set the learning rate as 0.05; for ResNet, we use 0.1. We respectively use a warm start with 100 training steps and 10 training steps for centralized training and decentralized training (during local training). We use batch size 10 for FeMnist and 64 for other datasets.
	
	We use FedAvg~\cite{FedAvg}, FedProx~\cite{FedProx}, FedOpt~\cite{FedOpt}, Scaffold~\cite{Scaffold}, and MOON~\cite{MOON} as base FL algorithms. For all of these algorithms, we take $H$ communication rounds, and select $R * 100.0\%$ clients during each round. Each client updates the global model on their private data for $E$ epochs. For FedProx, the regularization coefficient of the proximal term is tuned in $\{1e-4, 1e-3\}$ and the best one is reported. For FedOpt, we take SGD with momentum 0.9 as the global optimizer, and tune the global learning rate in $\{0.1, 0.5, 0.9\}$, which is similar to FedAvgM~\cite{FedAvgM}. We also try using Adam as the global optimizer and find the performances are not stable. For Scaffold, we use the implementation from the online page~\footnote{\url{https://github.com/ramshi236/Accelerated-Federated-Learning-Over-MAC-in-Heterogeneous-Networks}}. For MOON, we set the coefficient of the contrastive loss as $1.0$, which is recommended by the authors. We then replace the normal neurons with the proposed PANs to improve these algorithms. We keep $T=1$ by default and tune hyper-parameters from: PAN$_+$ with $A=0.05$, PAN$_\circ$ with $A=0.05$, PAN$_\circ$ with $A=0.1$.
	
	\begin{figure}
		\centering
		\includegraphics[width=\linewidth]{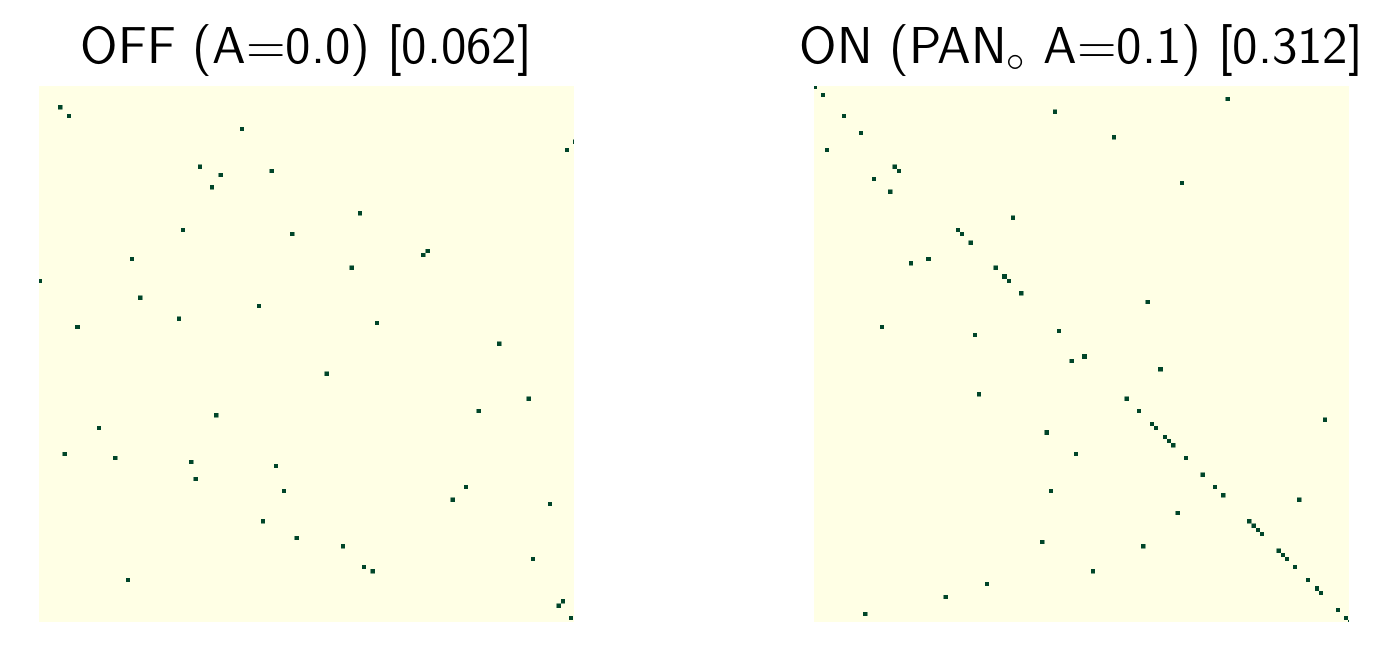}
		\caption{Optimal assignment matrix with PANs off/on, left vs. right. ({\scriptsize $\alpha=1.0$, $E=20$, VGG9 Conv6 on Cifar10.})}
		\label{fig:assign-vgg-conv6}
	\end{figure}
	
	\begin{figure}
		\centering
		\includegraphics[width=\linewidth]{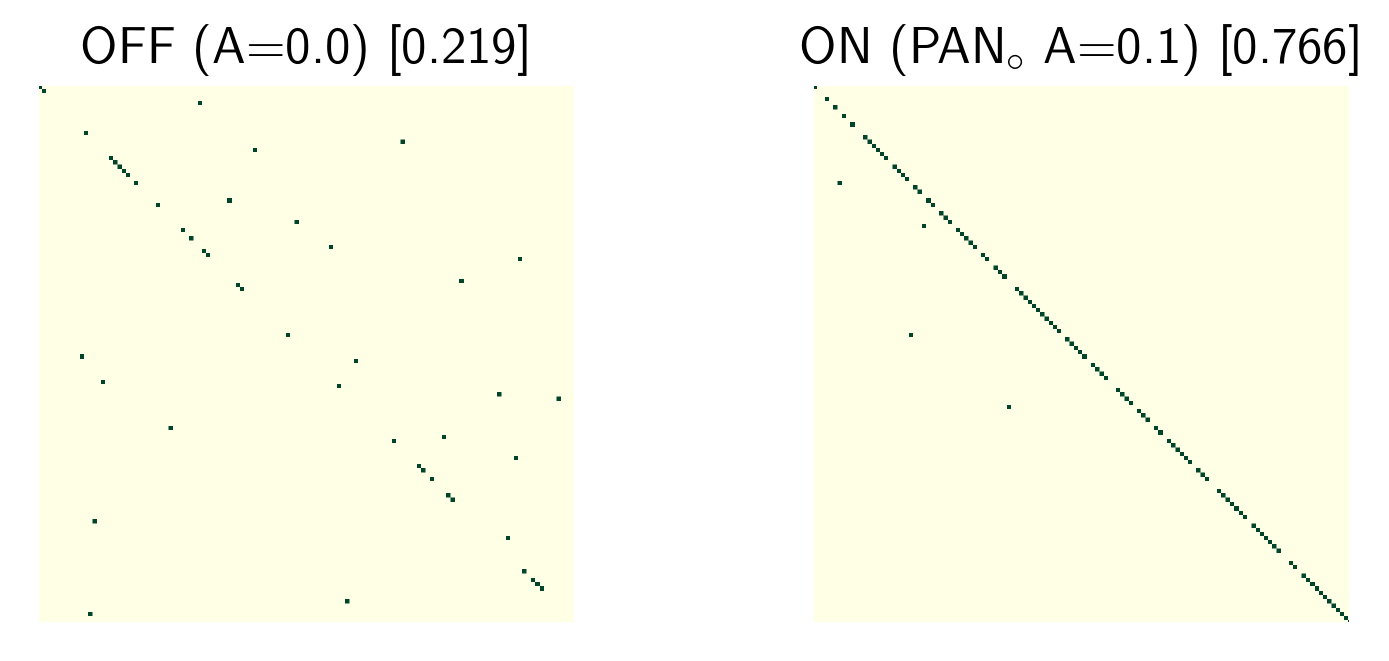}
		\caption{Optimal assignment matrix with PANs off/on, left vs. right. ({\scriptsize $\alpha=1.0$, $E=20$, MLP FC3 on Mnist.})}
		\label{fig:assign-mlp-fc3}
	\end{figure}
	
	\begin{figure}
		\centering
		\includegraphics[width=\linewidth]{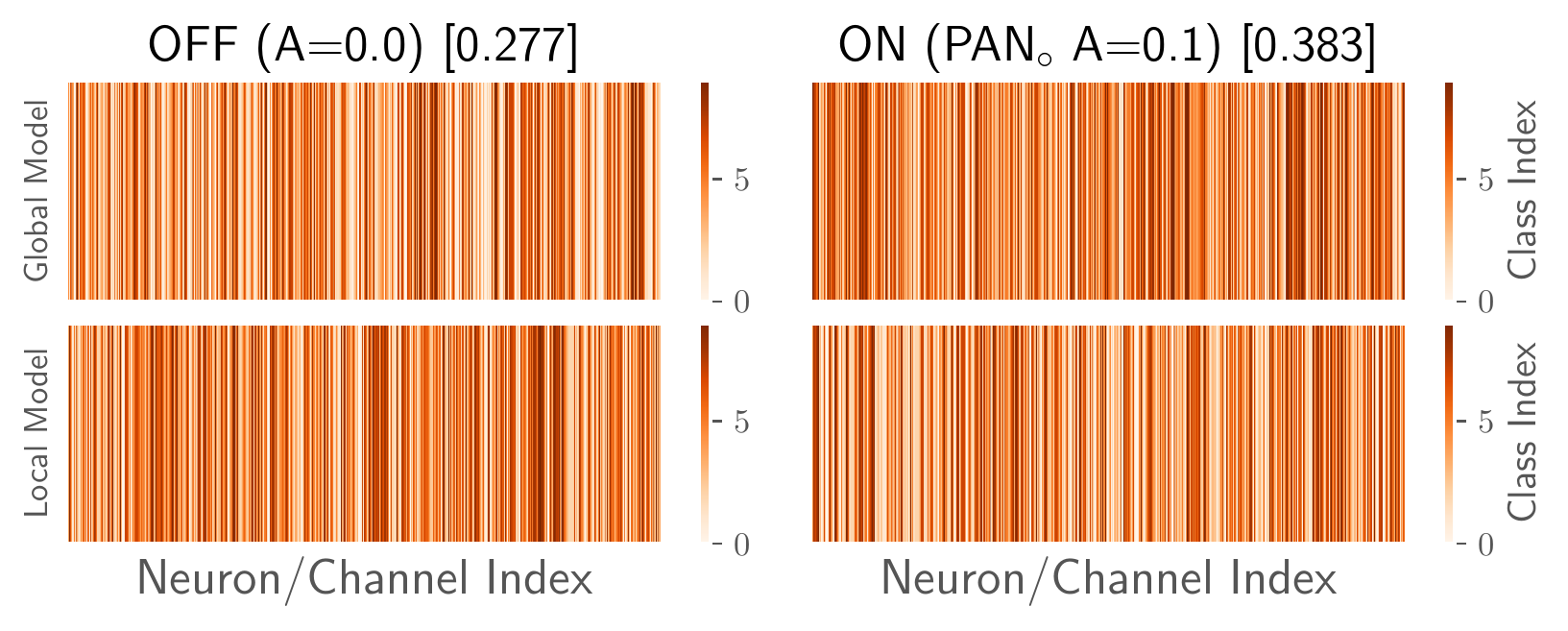}
		\caption{Preference vectors with PANs off/on, left vs. right. ({\scriptsize $\alpha=1.0$, VGG9 Conv5 on Cifar10.})}
		\label{fig:prefer-vgg-conv5}
	\end{figure}
	
	\begin{figure}
		\centering
		\includegraphics[width=\linewidth]{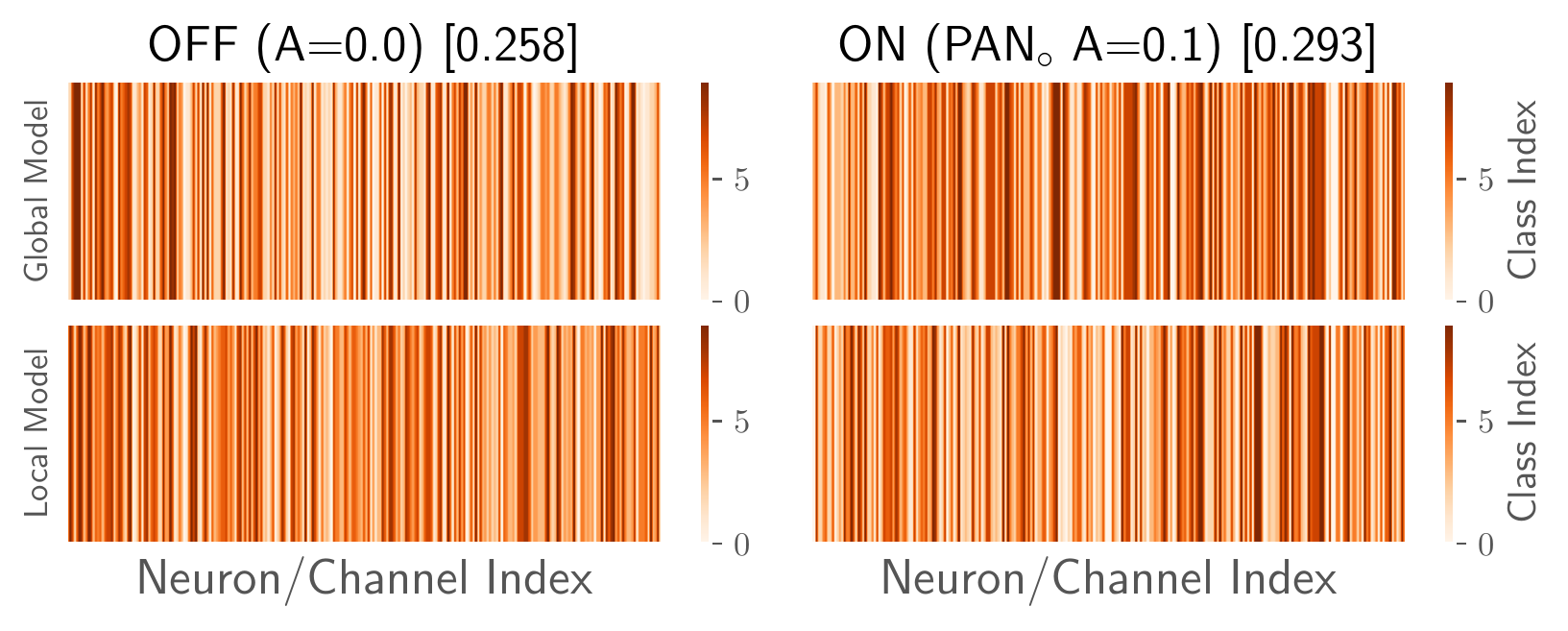}
		\caption{Preference vectors with PANs off/on, left vs. right. ({\scriptsize $\alpha=1.0$, VGG9 Conv4 on Cifar10.})}
		\label{fig:prefer-vgg-conv4}
	\end{figure}
	
	\begin{figure}
		\centering
		\includegraphics[width=\linewidth]{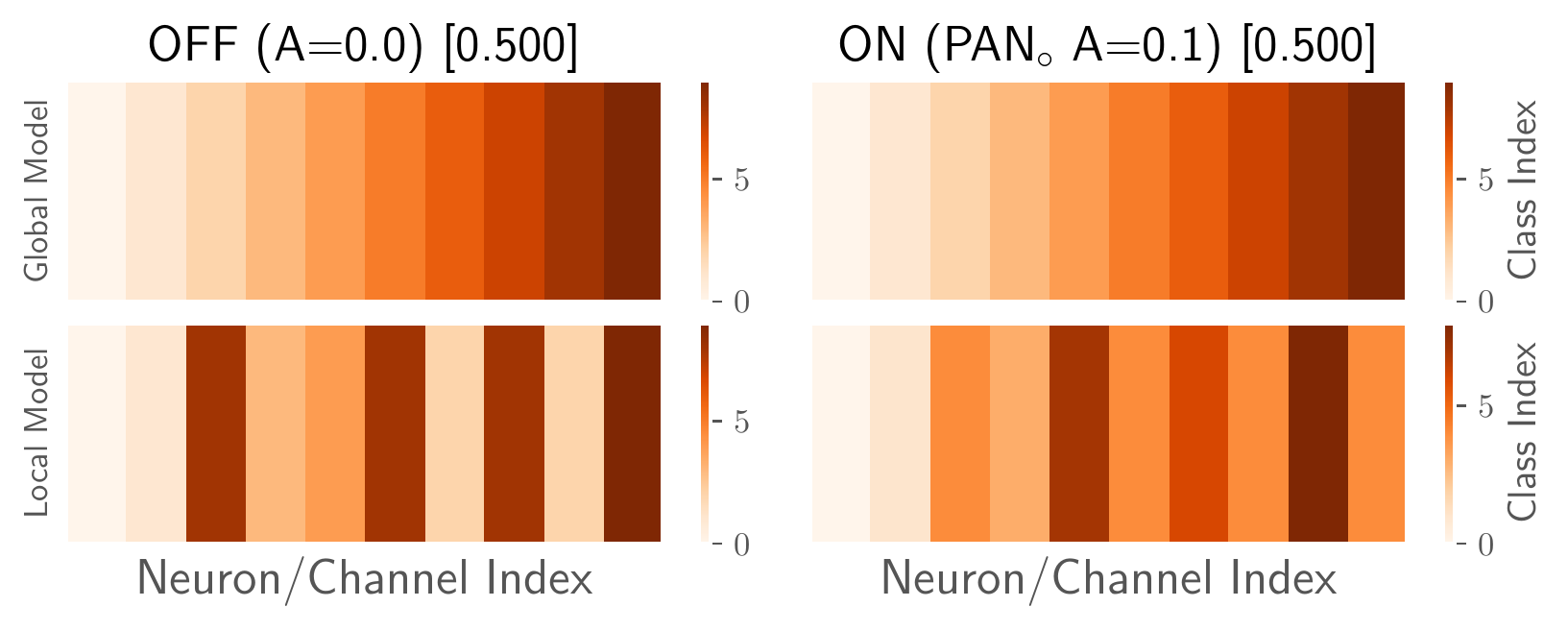}
		\caption{Preference vectors with PANs off/on, left vs. right. ({\scriptsize $\alpha=1.0$, MLP FC4 on Mnist.})}
		\label{fig:prefer-mlp-fc4}
	\end{figure}
	
	\begin{figure}
		\centering
		\includegraphics[width=\linewidth]{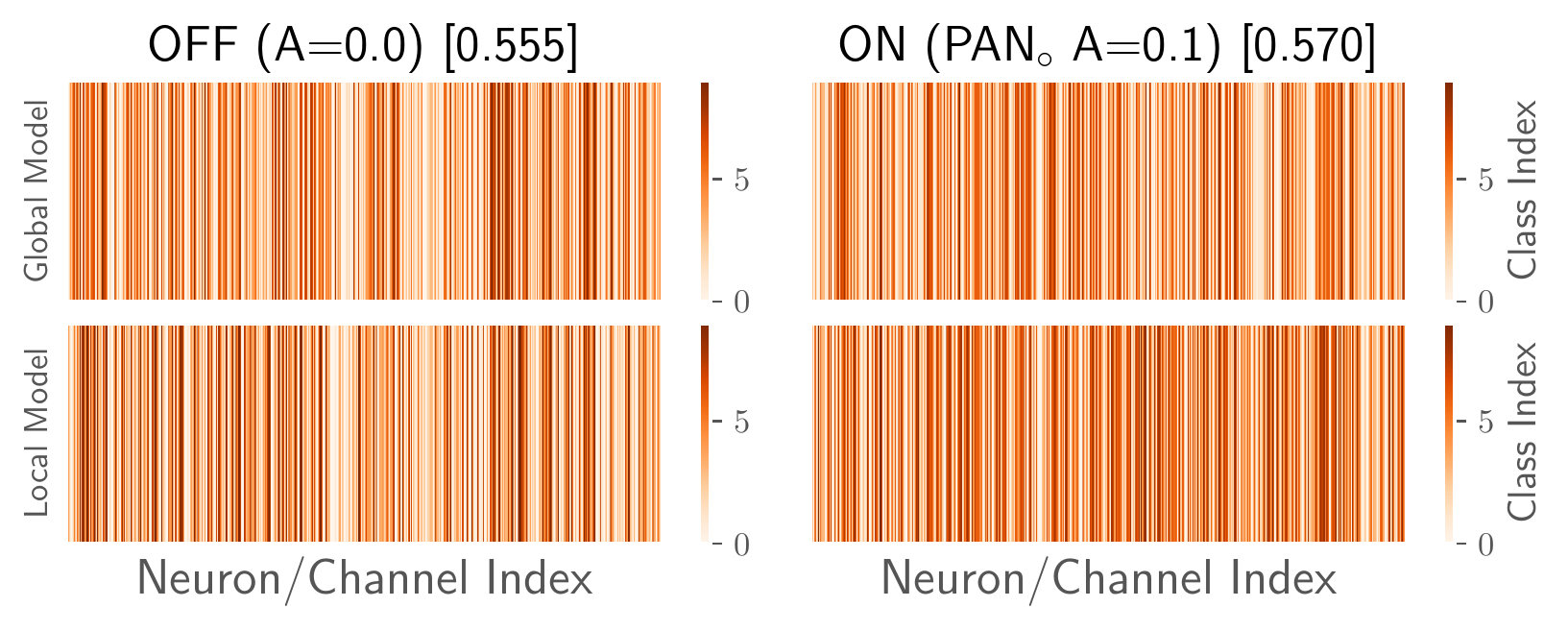}
		\caption{Preference vectors with PANs off/on, left vs. right. ({\scriptsize $\alpha=1.0$, MLP FC3 on Mnist.})}
		\label{fig:prefer-mlp-fc3}
	\end{figure}

	\begin{figure}
		\centering
		\includegraphics[width=\linewidth]{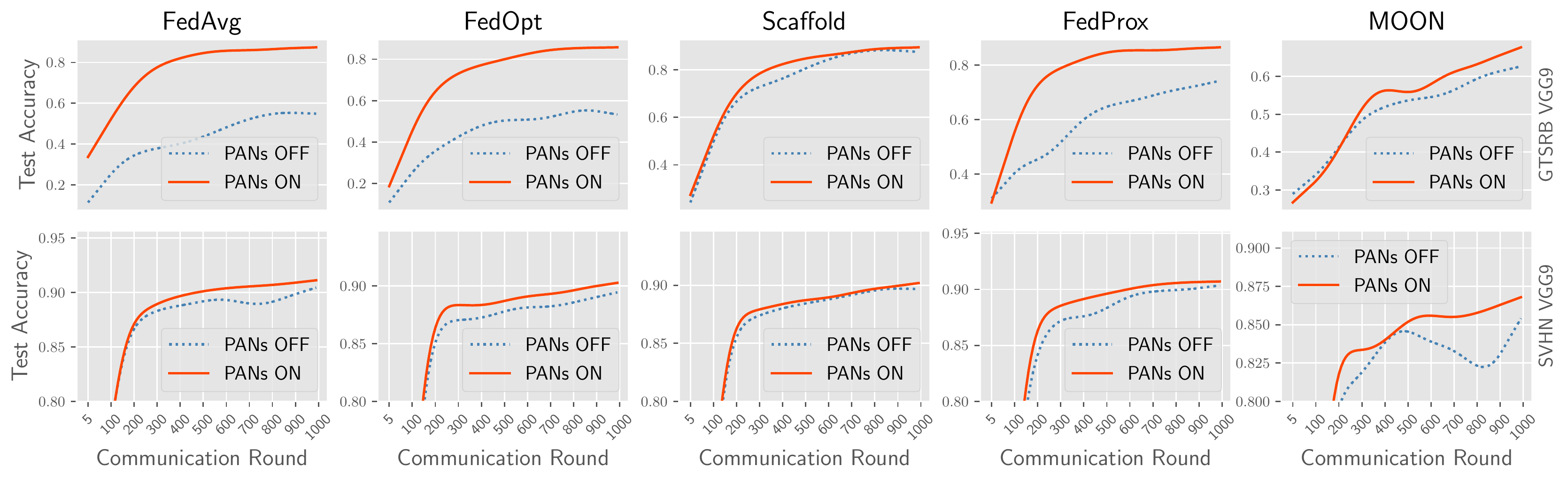}
		\caption{Comparison results on non-i.i.d. data ($\alpha$=0.1). Rows show datasets and columns show FL algorithms. PANs could universally improve these algorithms.}
		\label{fig:compare-noniid-supp}
	\end{figure}
	
	\begin{figure}
		\centering
		\includegraphics[width=\linewidth]{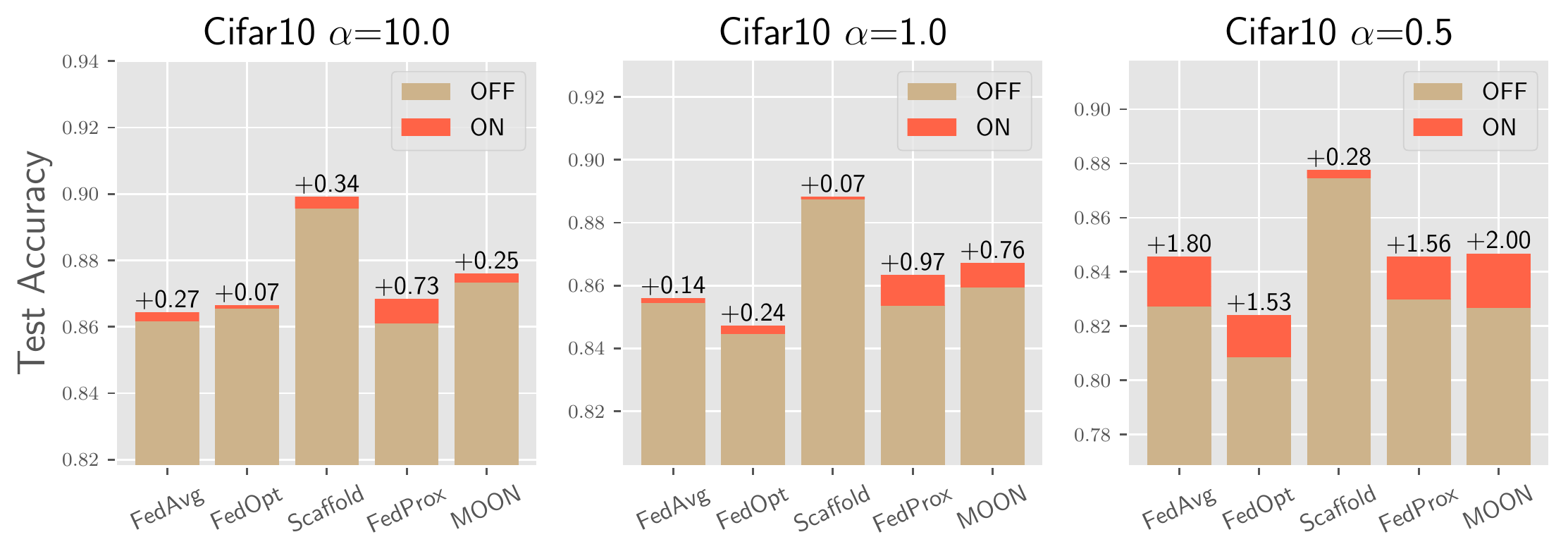}
		\caption{Comparisons under various levels of non-i.i.d. data on Cifar10. Smaller $\alpha$ implies more non-i.i.d. data.}
		\label{fig:compare-alpha-cifar10}
	\end{figure}
	
	\begin{figure}
		\centering
		\includegraphics[width=\linewidth]{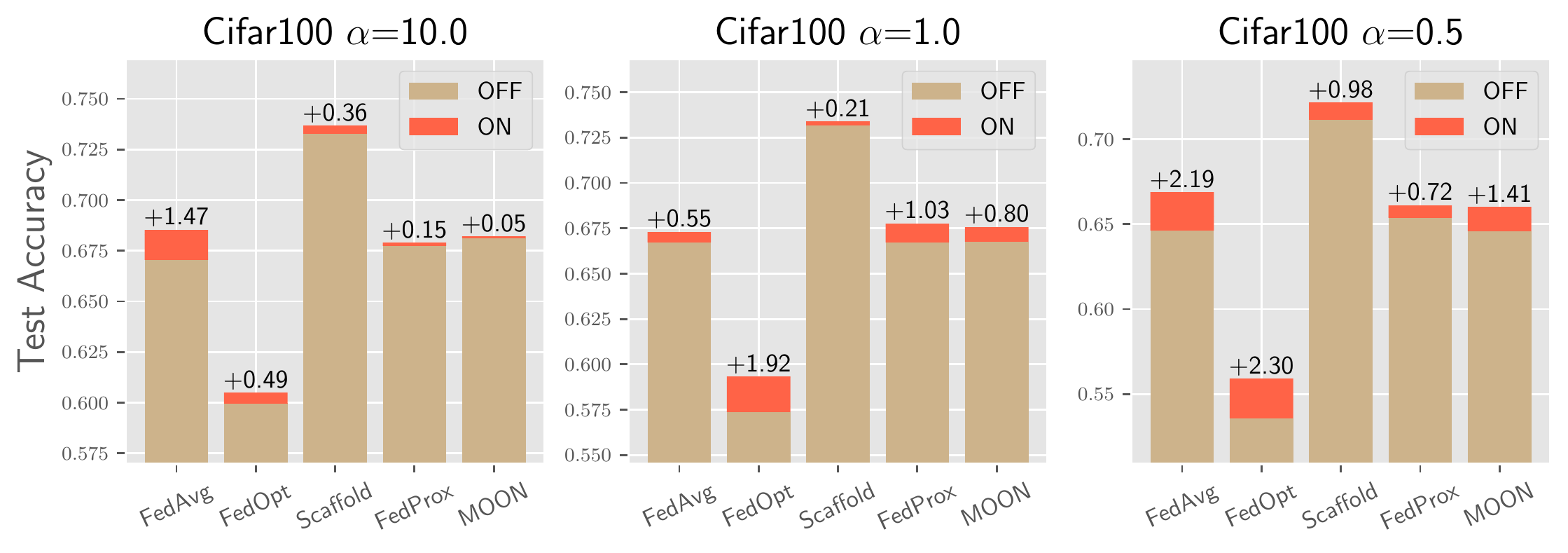}
		\caption{Comparisons under various levels of non-i.i.d. data on Cifar100. Smaller $\alpha$ implies more non-i.i.d. data.}
		\label{fig:compare-alpha-cifar100}
	\end{figure}
	
	\begin{figure}
		\centering
		\includegraphics[width=\linewidth]{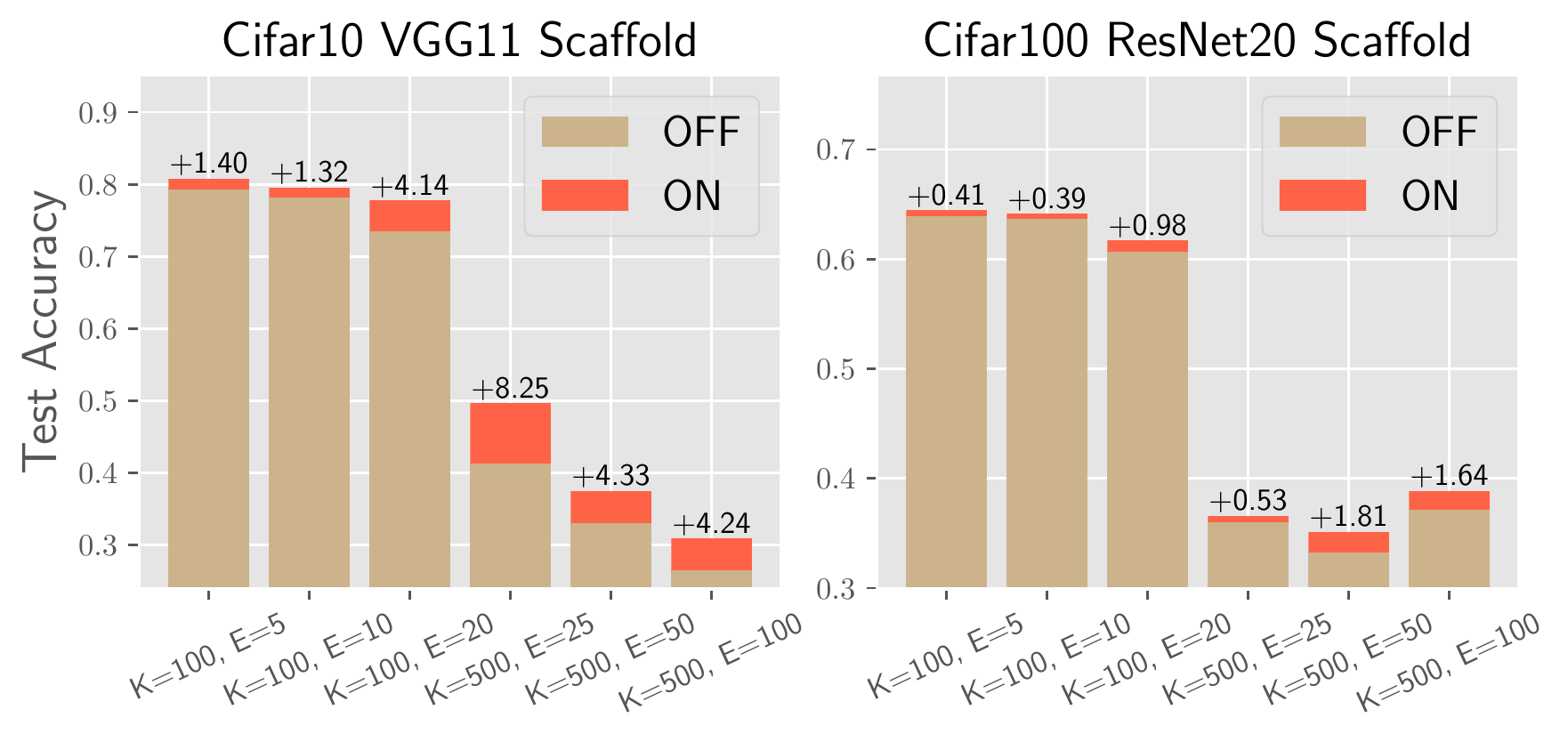}
		\caption{Comparisons under different FL scenes ($K$, $E$) based on Scaffold.}
		\label{fig:compare-scenes-scaffold}
	\end{figure}
	
	\begin{figure*}
		\centering
		\includegraphics[width=\linewidth]{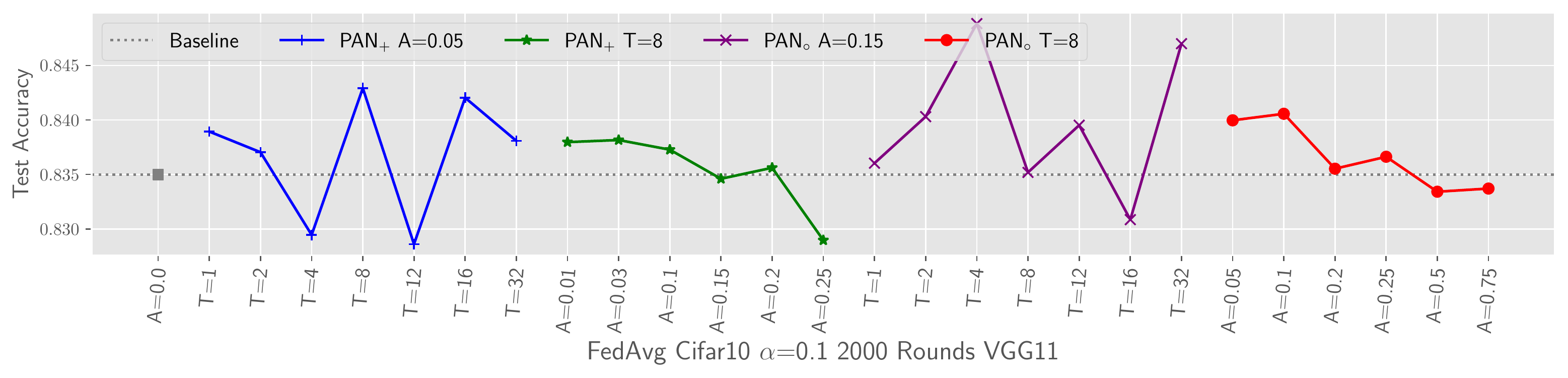}
		\caption{Hyper-parameter analysis on Cifar10 with VGG11.}
		\label{fig:hyper-cifar10}
	\end{figure*}
	
	\begin{figure*}
		\centering
		\includegraphics[width=\linewidth]{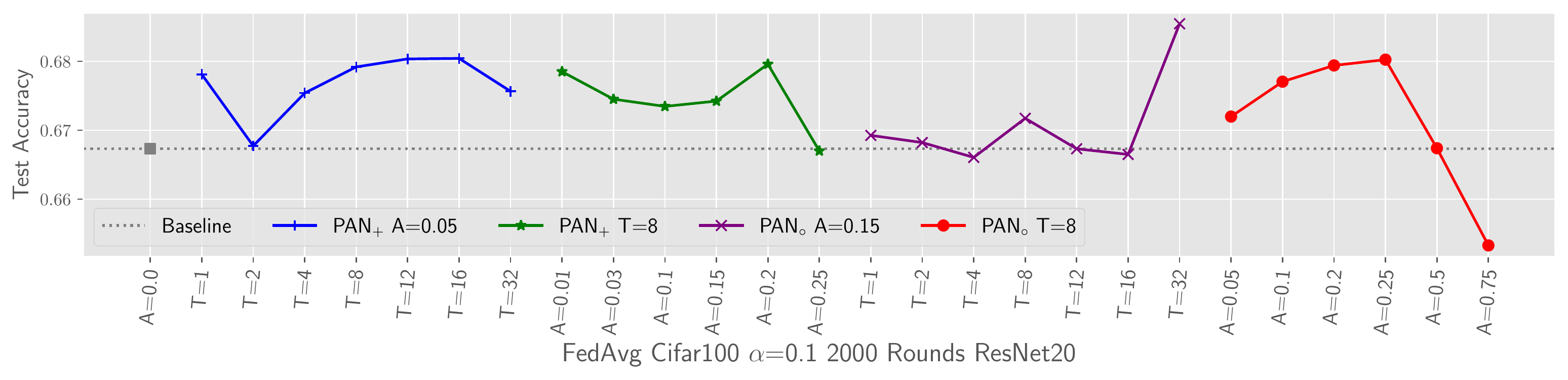}
		\caption{Hyper-parameter analysis on Cifar100 with ResNet20.}
		\label{fig:hyper-cifar100}
	\end{figure*}

	\begin{table*}
		\centering
		\begin{tabular}{@{}ccccccc@{}}
			\toprule
			& FeMnist & GTSRB & SVHN & Cifar10 & Cifar100 & Cinic10  \\
			& MLP & VGG9 & VGG9 & VGG11 & ResNet20 & ResNet20  \\
			\midrule
			SGD + Momentum=0.9 (LR in \{0.05,0.1\}) & 53.39 & 86.96 & 89.93 & 84.57 & 70.82 & 82.76 \\
			Adam (LR=3e-4) & 54.25 & 90.84 & 91.13 & 87.13 & 67.22 & 81.99 \\
			\bottomrule
		\end{tabular}
		\caption{The performances of centralized training with corresponding networks (without PANs), i.e., the \textbf{upper bound of decentralized training} (FL).}
		\label{tab:cent-acc}
	\end{table*}
	
	\begin{figure*}
		\centering
		\includegraphics[width=\linewidth]{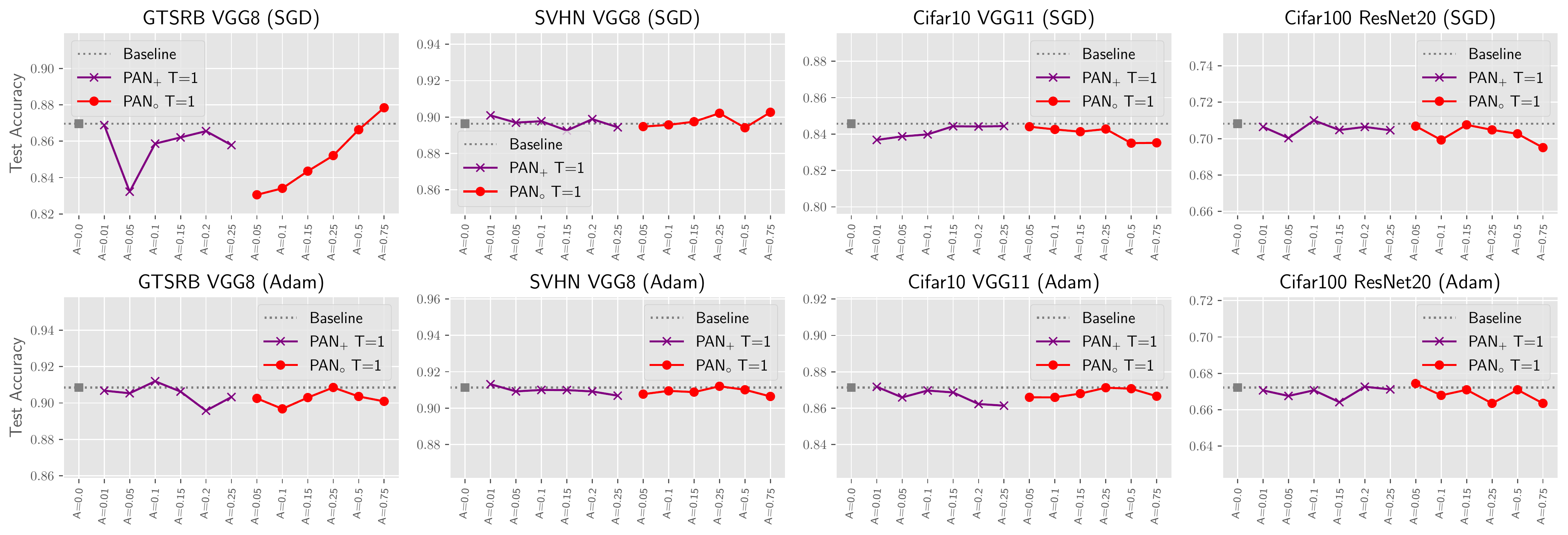}
		\caption{Performances of centralized training with PANs. The two parts respectively show the results of additive PANs and multiplicative PANs.}
		\label{fig:cent}
	\end{figure*}
	
	\section{Experimental Details}
	
	\subsection{Shuffle Test and Shuffle Test in FL}
	We propose a procedure to measure the degree of permutation invariance of a certain neural network, that is, how large the shuffle error is after shuffling the neurons. The shuffle process is shown in Alg.~\ref{algo:sf}, where $P_{\text{sf}}$ controls the disorder level of the constructed permutation matrices. Some additional descriptions are: (1) the permutation matrix (PM) should be randomly generated and we don't need to solve it; (2) PMs are introduced just to verify the property of PANs that they can disable the permutation invariance of neural networks, which is not used in our FedPAN algorithm; (3) the computational complexity is $\mathcal{O}(J)$, requiring at most J swaps, which is very efficient to implement during simulation.
	
	We introduce the shuffle test in the body of this paper. Specifically, we manually shuffle the network and study the output change, i.e., the shuffle error defined in the body. A hyper-parameter $P_{\text{sf}}$ is used to control the disorder of permutation. Given a $P_{\text{sf}}$, we could generate a permutation matrix $\Pi$, then we calculate how many neurons are not shuffled via computing ``$R_{\text{kept}}$=np.mean(np.diag($\Pi$))". We use the functions provided in the Numpy~\footnote{\url{https://numpy.org/}} package. This is calculated and its correspondence to $P_{\text{sf}}$ is shown in the body.
	The shuffle process is also applied to FL. Specifically, we present the Pseudo-Code in Alg.~\ref{algo:sf-fl}. Easily, the model will be shuffled for $N_{\text{sf}}$ times during local training in expectation. Hence, we calculate the corresponding $R_{\text{kept}}$ as the diagonal ones after several accumulative permutation, i.e., ``$R_{\text{kept}}$=np.mean(np.diag($\Pi_{r_k}\cdots\Pi_{2}\Pi_{1}$))", where $\Pi_{1}$, $\Pi_2$, and $\Pi_{r_k}$ denote the generated permutation matrices in each local update step. We simulate the process for a single layer 10 times and calculate the averaged $R_{\text{kept}}$. We keep $P_{\text{sf}}=0.1$ and show the relations of $R_{\text{kept}}$ and $N_{\text{sf}}$ in the body.

	\section{Additional Experimental Results}
	
	\noindent \textbf{Shuffle Error on Random Data:} We investigate the shuffle error via taking the random data as input in the body, where we only present the results based on VGG13. We report similar results on MLP and ResNet20, which are shown in Fig.~\ref{fig:sf-func-mnist} and Fig.~\ref{fig:sf-func-cifar}. Multiplicative PANs with a larger A make the network more sensitive to neuron permutation.
	
	\noindent \textbf{Weight Divergence:} Our proposed PANs could decrease the weight divergence during FL. Specifically, we split the training data onto $K=10$ clients with $\alpha \in \{1.0, 0.1\}$ and select all clients in each round, i.e., $R=1.0$. We take $H=20$ communication rounds and then calculate the local gradient variance as an approximation. We vary the number of local epochs $E \in \{5, 20\}$. We only report the results on Mnist with $E=5$ in the body. Additional results of Mnist with $E=20$ (Fig.~\ref{fig:wdiv-mnist-20}) and Cifar10 with $E=5$ (Fig.~\ref{fig:wdiv-cifar-5}) further verify that PANs could decrease the local gradient variance.
	
	\noindent \textbf{Matching via Optimal Assignment:} We first train a global model via FL for $H=20$ communication rounds, where the scene contains $10$ clients with $\alpha=1.0$. Then, we randomly sample a local client and update the global model for $E$ epochs. Our goal is to search for a matrix to match the neurons of the global model and the updated one, i.e., the local model of this client. We then use 500 test samples to obtain the neuron's activations as their representations. Hence, the optimal assignment problem could be solved and the assignment matrix is a permutation matrix. The results on various layers of VGG9 and MLP are shown in Fig.~\ref{fig:assign-vgg-conv6} and Fig.~\ref{fig:assign-mlp-fc3}. {\it Notably, the calculated matching ratio, i.e., the number in ``[]", is only an approximated value which represents how much neurons are shuffled. The absolute value (e.g., 0.062) does not represent the actual permutation during training.}
	
	\noindent \textbf{Visualizing Neurons via Preference Vectors:} Similarly, more of the visualization results via preference vectors of neurons are provided in Fig.~\ref{fig:prefer-vgg-conv5}, Fig.~\ref{fig:prefer-vgg-conv4}, Fig.~\ref{fig:prefer-mlp-fc4}, and Fig.~\ref{fig:prefer-mlp-fc3}. Notably, there are only 10 neurons in Fig.~\ref{fig:prefer-mlp-fc4} because FC4 is the output layer with 10 classes. Using PANs could encourage neurons at the same position contribute to the same classes as much as possible.

	\noindent \textbf{Universal Application of PANs:} We report the results of applying PANs to popular FL algorithms on FeMnist, Cifar10, Cifar100, and Cinic10 in the body. We show the results on SVHN and GTSRB in Fig.~\ref{fig:compare-noniid-supp}. Training on GTSRB is not stable, and some algorithms will converge slower, e.g., FedAvg and FedOpt. This could be improved with the additional effort of tuning learning rates, while we omit this in this paper. Comparison results on Cifar10 and Cifar100 under various levels of non-i.i.d. data are shown in Fig.~\ref{fig:compare-alpha-cifar10} and Fig.~\ref{fig:compare-alpha-cifar100}. The improvements under various scenes based on Scaffold are shown in Fig.~\ref{fig:compare-scenes-scaffold}. These additional results further verify the universal application of PANs to improve the performance of FL.

	\noindent \textbf{Hyper-parameter Analysis:} We present the performances of various $A$ with PAN$_{\circ}$ when $T=1$ in the body and point out that setting $A=0.1$ is a good choice. Here, we present a more comprehensive analysis with both additive and multiplicative PANs. The used FL scene is: $K=100$, $\alpha=0.1$, $H=2000$, $R=0.1$, $E=5$. We plot the results on Cifar10 with VGG11 and Cifar100 with ResNet20 in Fig~\ref{fig:hyper-cifar10} and Fig.~\ref{fig:hyper-cifar100}. The leftmost point shows the baseline of the performance. The four parts in different colors show the results with various $T$ or $A$, while the other one is fixed. For example, the first part shows the performances with $T \in \{1, 2, 4, 8, 12, 16, 32\}$ in PAN$_+$, while $A$ is fixed to 0.05. Clearly, with fixed $T$, a larger $A$ leads to degradation (the green and the red part). Setting $A$ around 0.1 for PAN$_\circ$ is recommended. The results on Cifar100 are more invariant to $T$, although the performances fluctuate a lot on Cifar10. Many of these hyper-parameters could surpass the baseline.
	
	\begin{figure}
		\centering
		\begin{subfigure}{0.48\linewidth}
			\includegraphics[width=\linewidth]{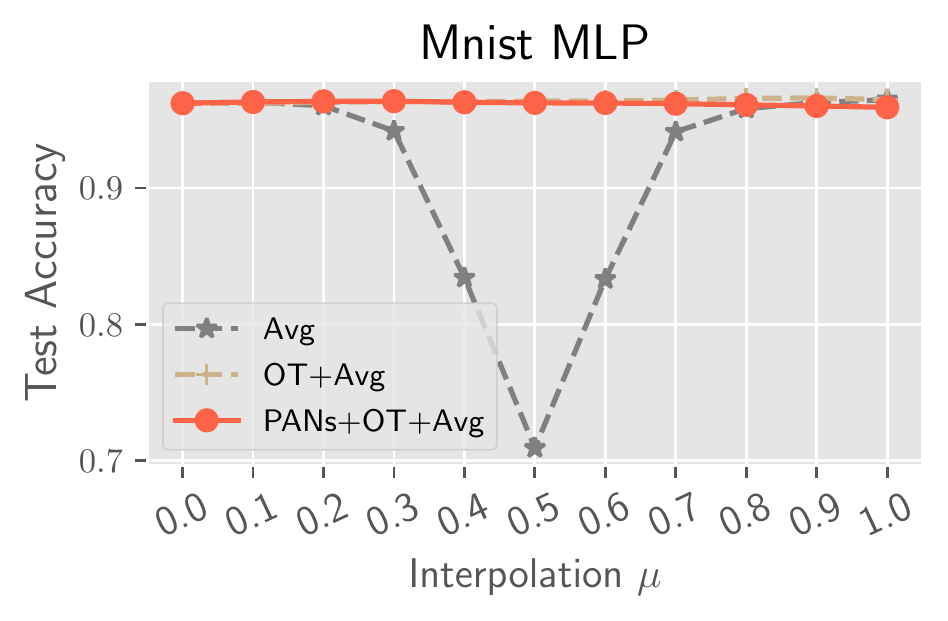}
		\end{subfigure}
		\hfill
		\begin{subfigure}{0.48\linewidth}
			\includegraphics[width=\linewidth]{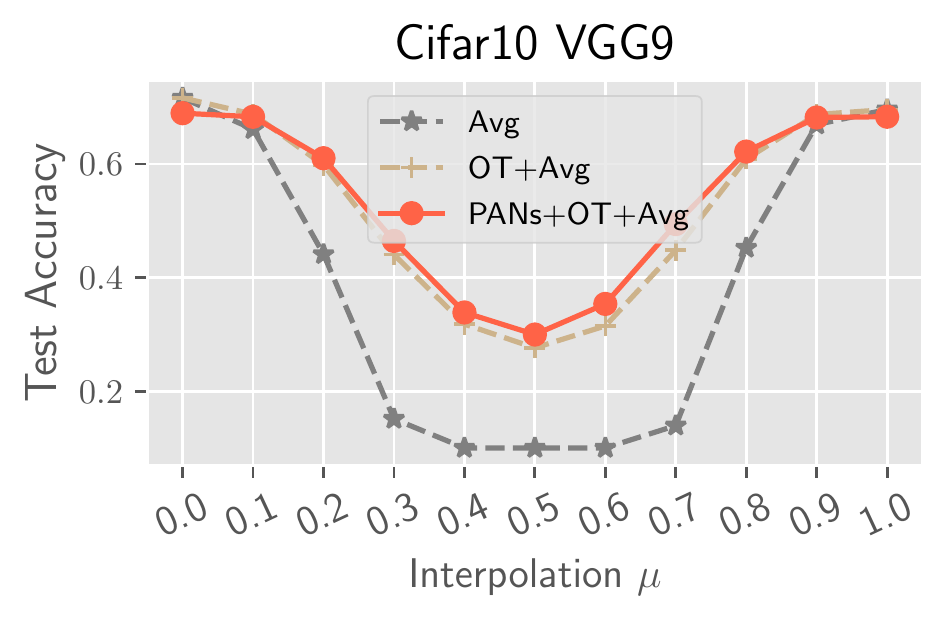}
		\end{subfigure}	
		\caption{Model fusion of MLP on Mnist (Left) and VGG9 on Cifar10 (Right) with direct parameter averaging, optimal transport, and PANs. The x-axis shows the interpolation coefficient.}
		\label{fig:ot}
	\end{figure}
	
	\begin{figure}
		\centering
		\includegraphics[width=\linewidth]{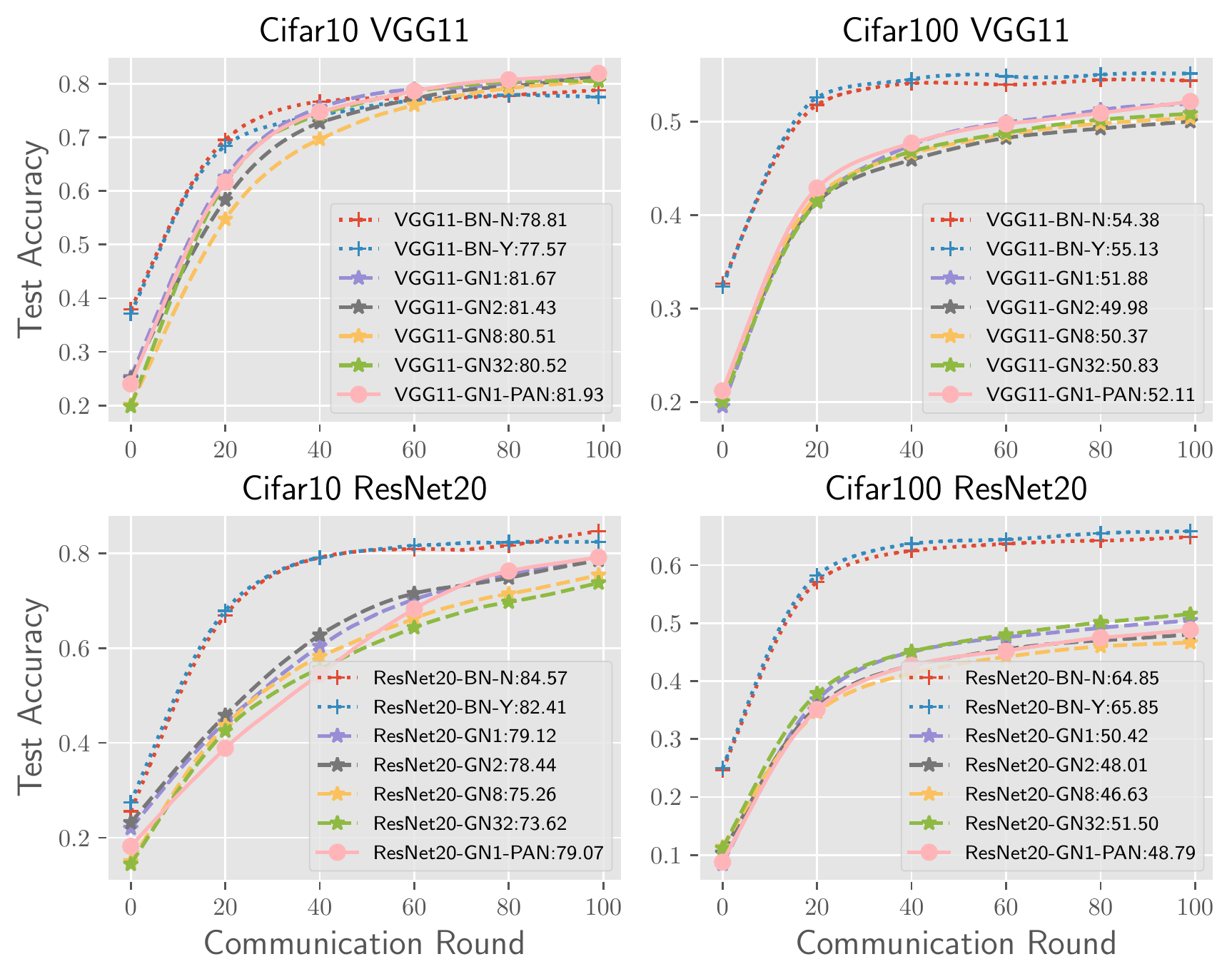}
		\caption{Comparisons of different normalization techniques in ConvNet. The top is based on VGG11 and the bottom is based on ResNet20. We use datasets Cifar10 and Cifar100.}
		\label{fig:compare-gn}
	\end{figure}

	\section{More Studies}
	\subsection{Centralized Training}
	We report the test accuracies of centralized training on FeMnist, GTSRB, SVHN, Cifar10, Cifar100, and Cinic10. The utilized networks are correspondingly MLP, VGG9, VGG9, VGG11, ResNet20, and ResNet20. The numbers of training epochs are respectively 30, 20, 30, 30, 100, and 100. We utilize both SGD with momentum 0.9 and Adam as the optimizer. For SGD, we use 0.05 as the learning rate for MLP and VGG, while 0.1 for ResNet20. For Adam, we use 0.0003 for all networks. The performances are listed in Tab.~\ref{tab:cent-acc}.
	We then add PANs to some datasets and find that the performances degrade slightly. We vary the hyper-parameter $A$ in PANs while keeping $T=1$. The results are shown in Fig.~\ref{fig:cent}. Using PANs could harm the training process slightly, and commonly, a larger $A$ could make the results worse. Although we try utilizing the adaptive optimizer (i.e., Adam), the results of utilizing PANs do not improve. Advanced optimizers should be proposed to mitigate the degradation, which is left for future work.
	
	\subsection{Optimal Transport for Model Fusion}
	FL should send down the global model to local clients as the initialization during each communication round. If not, coordinate-based parameter averaging will become worse. The work~\cite{OTFusion} studies model fusion with different initializations, and utilizes optimal transport~\cite{Barycenter} to align model parameters. We split Mnist and Cifar10 into two parts uniformly. We train independent models on these two sets correspondingly. The obtained models after training $20$ epochs are denoted as $\theta_A$ and $\theta_B$. Then, an interpolation is evaluated, i.e., $(1-\mu) \theta_A + \mu \theta_B$, $\mu \in [0,1]$. Directly averaging these two models will perform poorly, which is shown in Fig~\ref{fig:ot} (the line with legend ``Avg"). If we align the models via optimal transport and then interpolate the aligned models, the results become better (the line with legend ``OT+Avg" in Fig~\ref{fig:ot}). We further add PANs during model training and the performances could be slightly improved (the line with legend ``PANs+OT+Avg" in Fig~\ref{fig:ot}). This shows that PANs may still be helpful with different initializations.

	\subsection{BatchNorm vs. GroupNorm}
	We then investigate the normalization techniques in deep neural networks. Previous FL works point out that GroupNorm may be more applicable to FL with non-iid data~\cite{NonIID-Quag}. Specifically, BatchNorm calculates the mean and variance of a data batch, which is relevant to local training data. Hence, the statistical information in BatchNorm will diverge a lot across clients. One solution is aggregating the statistical information during FL, i.e., averaging the ``running mean" and ``running variance" in BatchNorm. We denote this as ``BN-Y". In contrast, we use ``BY-N" to represent the method that ``running mean" and ``running variance" are not aggregated. We also vary the number of groups in GroupNorm, i.e., $\{1,2,8,32\}$, which are denoted as ``GN1", ``GN2", ``GN8", and ``GN32". We list the convergence curves on Cifar10 and Cifar100 in Fig.~\ref{fig:compare-gn}. We use VGG11 and ResNet20 as the backbone. The numbers in the legends denote the final test accuracies. GroupNorm only improves the performances of Cifar10 with VGG11. Additionally, setting the number of groups as 1 is better. We also apply PANs to networks with ``GN1" and find the performance does not improve. The combination of PANs with various normalization techniques is also interesting, which is also left for future work.
	
	\subsection{Personalization in FL}
	Finally, we present some possible varieties of PANs for personalization in FL. In the body of this paper, we take the same position encodings among clients and implicitly make neurons combined with their positions. However, if we take different position encodings or partially shared position encodings among clients, we could let similar clients contribute more. Some clients own individual positions, which could be utilized for personalization. These ideas are also left for future work.

\end{document}